%% file: arxiv.tex
\PassOptionsToPackage{dvipsnames}{xcolor}
\documentclass[11pt,letterpaper]{style}
\usepackage[numbers,sort&compress]{natbib}
\usepackage{graphicx}
\usepackage{booktabs}
\usepackage{amsmath,amsfonts,amssymb}
\usepackage{cleveref}
\usepackage{wrapfig}
\usepackage{multirow}
\usepackage{colortbl}
\usepackage{listings}
\usepackage{xparse}
\usepackage{fontawesome5}
\usepackage{bxcoloremoji}
\usepackage{float}
\usepackage{placeins}
\usepackage{threeparttable}

\graphicspath{{./}{fig/}{figures/}{plot/}{pdf/}{table/}}
\usepackage{amsthm}
\newtheorem{fact}{Stylized Fact}    
\usepackage{tcolorbox}
\usepackage{algorithm}
\usepackage{algpseudocode}
\usepackage{svg} 
\usepackage{subcaption}

\tcbuselibrary{skins, breakable}

\renewenvironment{fact}[1][]{
    \refstepcounter{fact}
    \begin{tcolorbox}[
        colback=lightyellow,
        colframe=black!60,
        boxrule=0.6pt,
        arc=3mm,
        left=6pt, right=6pt, top=5pt, bottom=5pt,
        before skip=6pt, after skip=6pt,
        breakable
    ]
    \textit{\textbf{Stylized Fact \thefact.}}~
}{
    \end{tcolorbox}
}
\renewcommand{\thefact}{\Roman{fact}}


\tcbuselibrary{listingsutf8} 
\usepackage{titletoc}
\NewDocumentEnvironment{minted}{O{} m +b}{%
}{}

\newcommand{\equal}{\textsuperscript{*}}
\newcommand{\internship}{\textsuperscript{\dag}}
\newcommand{\corresponding}{\textsuperscript{\ddag}}
\newcommand{\hficon}{\raisebox{-0.2em}{\includegraphics[height=1em]{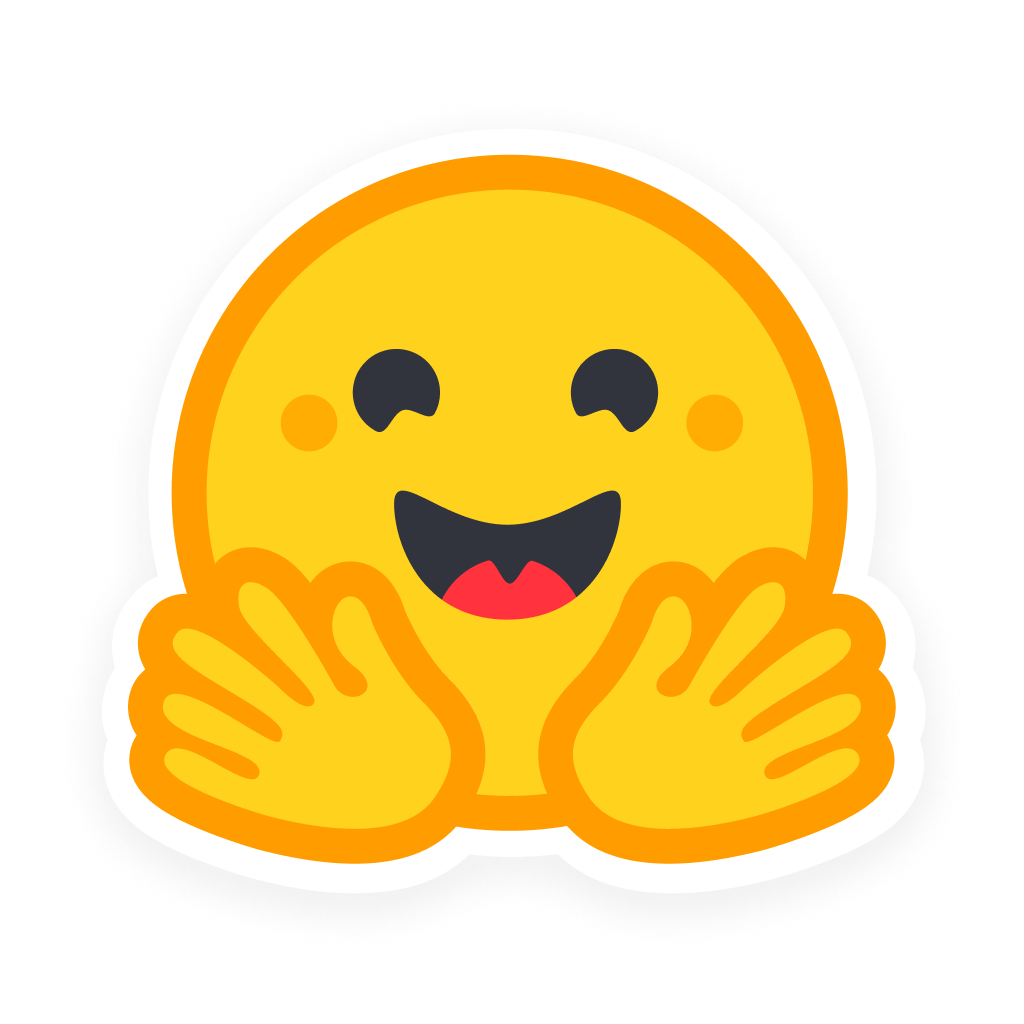}}}

\definecolor{myblue1}{HTML}{0171DC}
\definecolor{myblue2}{HTML}{013978} 
\title{Reasoning as Gradient: Scaling MLE Agents Beyond Tree Search}

\runningtitle{Reasoning as Gradient: Scaling MLE Agents Beyond Tree Search}

\renewcommand\Authfont{\centering\normalfont\bfseries\fontsize{11}{15}\selectfont}
\renewcommand\Affilfont{\centering\normalfont\fontsize{10}{15}\selectfont}

\author{%
    {\Authfont
    \textbf{Yifei Zhang}\textsuperscript{1}\equal\internship, \textbf{Xu Yang}\textsuperscript{2}\equal, \textbf{Xiao Yang}\textsuperscript{2}\equal, \\
    \textbf{Bowen Xian}\textsuperscript{2}, \textbf{Qizheng Li}\textsuperscript{3}\internship, \textbf{Shikai Fang}\textsuperscript{4}, \textbf{Jingyuan Li}\textsuperscript{5}\internship, \\
    \textbf{Jian Wang}\textsuperscript{2}, \textbf{Mingrui Xu}\textsuperscript{6}\internship, \textbf{Yuge Zhang}\textsuperscript{2}, \textbf{Weiqing Liu}\textsuperscript{2}\corresponding, \textbf{Jiang Bian}\textsuperscript{2}
    }
    \\
    {\Affilfont
    \textsuperscript{1}Nanjing University, \textsuperscript{2}Microsoft Research Asia, \textsuperscript{3}Peking University,
    \\
    \textsuperscript{4}Zhejiang University, \textsuperscript{5}Wuhan University, \textsuperscript{6}Hong Kong University of Science and Technology
    \\
    \href{https://github.com/microsoft/RD-Agent}{\faGithub\ GitHub Repo} \quad
    \href{https://huggingface.co/datasets/amstrongzyf/Gome-GPT5-Traces}{\hficon\ GPT-5 Execution Traces}

    \vspace{8pt} }
}

\begin{document}
\begin{abstract}
\textbf{Abstract:} LLM-based agents for machine learning engineering (MLE) predominantly rely on tree search, a form of gradient-free optimization that uses scalar validation scores to rank candidates. As LLM reasoning capabilities improve, exhaustive enumeration becomes increasingly inefficient compared to directed updates, analogous to how accurate gradients enable efficient descent over random search.
We introduce \textsc{Gome}, an MLE agent that operationalizes gradient-based optimization. \textsc{Gome} maps structured diagnostic reasoning to gradient computation, success memory to momentum, and multi-trace execution to distributed optimization. Under a closed-world protocol that isolates architectural effects from external knowledge, \textsc{Gome} achieves a state-of-the-art 35.1\% any-medal rate on MLE-Bench with a restricted 12-hour budget on a single V100 GPU.
Scaling experiments across 10 models reveal a critical crossover: with weaker models, tree search retains advantages by compensating for unreliable reasoning through exhaustive exploration; as reasoning capability strengthens, gradient-based optimization progressively outperforms, with the gap widening at frontier-tier models. Given the rapid advancement of reasoning-oriented LLMs, this positions gradient-based optimization as an increasingly favorable paradigm. We release our codebase and GPT-5 traces at: \url{https://github.com/microsoft/RD-Agent}.
\end{abstract}

\maketitle
\begingroup
\renewcommand{\thefootnote}{\fnsymbol{footnote}}
\footnotetext[1]{Equal contribution.}
\footnotetext[2]{Work done during an internship at Microsoft Research Asia.}
\footnotetext[3]{Correspondence: \texttt{Weiqing.Liu@microsoft.com}}
\endgroup

\input{paper_body.tex}

\bibliographystyle{plainnat}
\bibliography{custom}

\clearpage
\appendix
\input{appendix_arxiv.tex} 
\end{document}

%% file: paper_body.tex
\section{Introduction}

Automating Machine Learning Engineering (MLE) remains a long-standing challenge. Given a dataset and an evaluation metric, the objective is to autonomously complete the end-to-end development pipeline, including data preprocessing, feature engineering, model selection, and hyperparameter tuning, to deliver a high-performing predictive system. With rapid advances in code generation and complex reasoning capabilities of large language models~\citep{jaech2024openai,guo2025deepseek}, LLM-based MLE agents have recently emerged~\citep{aide2025,liu2025ml,nam2025mle}, demonstrating the ability to interpret task descriptions, diagnose data characteristics, implement complete solutions, and iteratively refine them using execution feedback. To enable systematic comparison of these agents, MLE-Bench~\citep{chan2024mle-bench} provides a standardized benchmark of 75 Kaggle competitions.

Current MLE agents predominantly adopt search-based exploration. AIDE~\citep{aide2025} first introduced this design motivated by two practical considerations: constrained context windows that made chain-based prompting prone to prompt bloat, and multi-branch exploration to mitigate uncertainty in single-shot generations. Subsequent work largely follows this paradigm: ML-Master~\citep{liu2025ml} incorporates Monte Carlo Tree Search, and AIRA~\citep{toledo2025ai} proposes graph-based exploration. Importantly, these methods use \emph{directed} topologies (e.g., UCT/PUCT-style node selection with value backpropagation). Our characterization of ``gradient-free'' is therefore about the \emph{optimization signal}, not the search topology: improvement is still primarily selected by scalar rewards/scores rather than by explicitly reasoning about update direction from rich execution feedback. Under this signal-level view, prior search-based agents exhibit two limitations. First, they are driven by \textbf{score-centric selection}: rich execution feedback (error traces, training dynamics, detailed logs) is compressed into scalar rewards that mainly decide which node/branch survives, discarding diagnostic information needed to determine how to update. This information loss becomes increasingly costly as LLM reasoning capabilities improve, since stronger models could otherwise extract precise improvement directions from such feedback. Second, these methods operate over a \textbf{predefined action space}: the agent selects among fixed templates, which cannot capture the effectively continuous nature of code modifications and may mismatch the specific failure modes revealed by execution.

\begin{figure}[t]
  \centering
  \includegraphics[width=0.55\columnwidth]{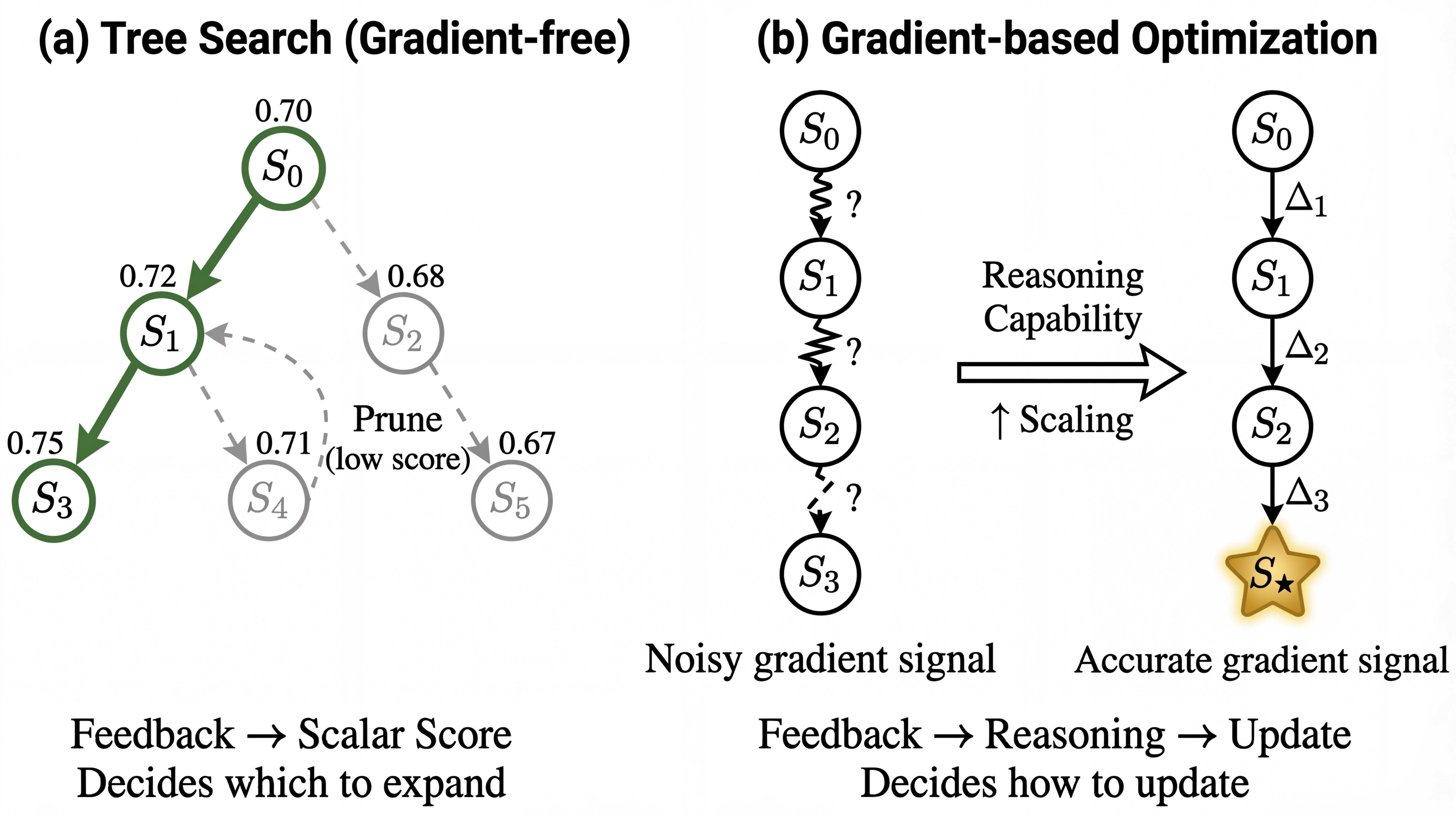}

  \caption{Two paradigms for utilizing execution feedback. (a)~Score-centric: feedback is compressed into scalar rewards that guide node/branch selection (which may itself be directed, e.g., via UCT/PUCT). (b)~Gradient-based: structured reasoning over execution feedback determines how to update the current solution; stronger reasoning yields more accurate signals.}
  \label{fig:paradigm}
\end{figure}

We argue that MLE inherently favors gradient-based optimization. Unlike domains punctuated by hard dead ends, MLE pipelines are typically repairable: improvements come from iteratively refining the same pipeline rather than discarding intermediate states. Moreover, the space of code modifications is effectively continuous, ranging from minor hyperparameter tweaks to architectural changes, favoring methods that propose state-conditioned updates based on execution feedback, rather than searching over a fixed action vocabulary. The key question is when such updates become preferable to enumeration. This trade-off is well understood in classical optimization: gradient-based methods dominate when gradient estimates are accurate, but degrade when noisy~\citep{maheswaranathan2019guided,asadi2024comparative}. In the MLE agent setting, the LLM's reasoning capability determines gradient signal quality. When reasoning is weak, multi-branch enumeration hedges against unreliable proposals; as reasoning-oriented models advance,\footnote{From o1-preview~\citep{jaech2024openai} to DeepSeek-R1~\citep{guo2025deepseek}, o3~\citep{openai_introducing_o3_2025}, and GPT-5~\citep{singh2025openai}.} the balance should shift toward gradient-based optimization: \textbf{gradient-based methods should increasingly outperform tree search as reasoning capability improves} (Figure~\ref{fig:paradigm}).

To test this claim, we propose \textbf{\textsc{Gome}} (\textbf{G}radient-based \textbf{O}ptimization for \textbf{M}achine Learning \textbf{E}ngineering), an agent that adopts gradient-based optimization in place of enumeration-driven search. \textsc{Gome} uses a chain-based iterative framework where each step updates the current solution along LLM-generated improvement directions, organized into modular components that mirror classical optimizer modules (Table~\ref{tab:mapping}). We evaluate under a \textbf{closed-world protocol} to isolate architectural effects from knowledge augmentation: agents use only task-provided materials and execution feedback, without external knowledge retrieval. Under this protocol, \textsc{Gome} achieves a \textbf{35.1\%} any-medal rate on \textsc{MLE-Bench} with a \textbf{12-hour} budget on a single\textbf{ V100} GPU, surpassing the previous state-of-the-art held by search-based methods. Component ablations confirm that each optimizer-inspired module contributes meaningfully. Scaling analysis, which compares \textsc{Gome} against tree-search baselines (including an MCTS variant sharing \textsc{Gome}'s codebase and previous state-of-the-art search-based methods) across 10 models ranging from non-reasoning (GPT-4o) to frontier reasoning models (GPT-5), further validates our claim: while tree-search methods show advantages with weaker reasoners, \textsc{Gome}'s advantage grows with model capability. Given the rapid advancement of reasoning-oriented LLMs~\citep{sun2025survey,xu2025towards}, this positions gradient-based optimization as an increasingly favorable paradigm.

In summary, this work makes the following key contributions: (1) We propose \textsc{Gome}, an MLE agent that adopts gradient-based optimization rather than score-centric candidate ranking, establishing a functional mapping between agent components and classical optimizer modules. (2) \textsc{Gome} achieves state-of-the-art performance on \textsc{MLE-Bench} under closed-world evaluation. Scaling analysis across 10 models validates our claim: \textsc{Gome}'s advantage over tree search grows with model reasoning capability, confirming that stronger reasoning enables more effective gradient-based optimization. (3) We release our codebase and GPT-5 execution traces to support reproducibility and contribute to the community.
\section{Related Work}

\paragraph{LLM-based MLE Agents.}
Current MLE agents explore various search topologies: tree search (AIDE~\citep{aide2025}, ML-Master~\citep{liu2025ml}, KompeteAI~\citep{kulibaba2025kompeteai}), graph-based frameworks (AIRA~\citep{toledo2025ai}, InternAgent-MLE~\citep{duinternagent}), and evolutionary methods (FM Agent~\citep{li2025fm}). Some approaches augment search with external knowledge retrieval from Kaggle notebooks, arXiv papers, or expert knowledge bases. MLE-STAR~\citep{nam2025mle} adopts a chain structure with web retrieval and targeted code-block refinement; however, its iteration remains score-driven, using ablation-based scoring to select which block to refine rather than reasoning about how to update (see Appendix~\ref{app:mle-star-comparison}). We summarize key design choices in Table~\ref{tab:comparison}. Notably, AIRA formalizes research agents as $(\mathcal{F}, \pi_{\text{sel}}, \mathcal{O}, \pi_{\text{op}}, \tau)$ and shows that, under the AIDE operator set, changing the search policy alone yields only limited gains, suggesting that operator design can be a major bottleneck beyond search topology~\citep{toledo2025ai}. InternAgent-MLE~\citep{duinternagent} further extends tree search to graph search with cross-branch trajectory reuse, yet the optimization signal remains score-driven. These examples illustrate a common pattern: topologies grow increasingly directed, but feedback utilization stays score-centric. In contrast, \textsc{Gome} integrates scalar scores with execution feedback for update: structured reasoning over these signals generates improvement hypotheses that directly guide solution modification.

\begin{table}[htbp]
  \centering
  \small
  \begin{tabular}{@{}lcccc@{}}
    \toprule
    \textbf{Method} & \textbf{Structure} & \textbf{Feedback Role} & \textbf{Ext. Know.} & \textbf{Code} \\
    \midrule
    AIDE & Tree & Ranking & \ding{55} & \ding{51} \\
    ML-Master & Tree & Ranking & \ding{55} & \ding{51} \\
    AIRA & Graph & Ranking & \ding{55} & \ding{51} \\
    MLE-STAR & Chain & Ranking & \ding{51} & \ding{51} \\
    InternAgent-MLE & Graph & Ranking & \ding{51} & \ding{55} \\
    KompeteAI & Tree & Ranking & \ding{51} & \ding{55} \\
    FM Agent & Evolutionary & Ranking & \ding{51} & \ding{55} \\
    \midrule
    \textsc{Gome} & Chain & Update & \ding{55} & \ding{51} \\
    \bottomrule
  \end{tabular}%
  \caption{Comparison of LLM-based MLE agents. Feedback Role indicates how execution feedback drives the primary optimization loop: Ranking denotes score-centric selection among candidates; Update denotes diagnostic reasoning over structured feedback to generate state-conditioned code modifications.}
  \label{tab:comparison}
\end{table}

\paragraph{Reasoning as Optimization.}
A growing line of work operationalizes LLM reasoning as an optimization signal, using task outcomes to iteratively refine artifacts rather than exhaustively enumerating candidates~\citep{yang2023large, shinn2023reflexion, zhang2025process}.
In prompt optimization, LLMs propose edits conditioned on performance feedback, with methods ranging from discrete optimization procedures~\citep{pryzant2023automatic, cui2024introducing} to population-based evolution~\citep{guo2023evoprompt}.
In agentic settings, trial outcomes are converted into self-reflections that guide behavior updates~\citep{madaan2023self}, while textual ``gradients'' formalize refinement as structured edits~\citep{yuksekgonul2024textgrad,cheng2024trace}.
While this paradigm has been widely adopted in prompt tuning and general-purpose agent tasks, it remains underexplored in MLE, where existing methods still rely on score-based search (Table~\ref{tab:comparison}).
\textsc{Gome} bridges this gap, instantiating reasoning-as-optimization for MLE with structured feedback as the gradient signal for code updates.

\section{\textsc{Gome}}
\label{sec:method}

We formulate MLE tasks as finding an optimal solution within the code space under resource constraints:
\begin{equation}
  s^* = \arg\max_{s \in \mathcal{S}} h(\mathcal{T}, s)
  \quad \text{s.t.} \quad \texttt{cost}(s) \le B,
\end{equation}
where $\mathcal{S}$ is the space of valid ML pipelines, $h(\mathcal{T}, s)$ evaluates solution $s$ on task $\mathcal{T}$ (e.g., accuracy), and $B$ bounds the total computational budget.

Existing MLE agents typically organize $\mathcal{S}$ as a tree or graph and use scalar scores to guide expansion. We propose \textsc{Gome}, which instead focuses on how to update. Inspired by gradient-based optimization, \textsc{Gome} extracts directional improvement signals from structured feedback via LLM reasoning. While true gradients are undefined in discrete code space, LLM reasoning serves a functionally analogous role: analyzing execution results to determine not just whether a solution improved, but why it improved and what to change next. Table~\ref{tab:mapping} summarizes this analogy.

\begin{figure}[ht]
  \centering
  \includegraphics[width=0.9\textwidth]{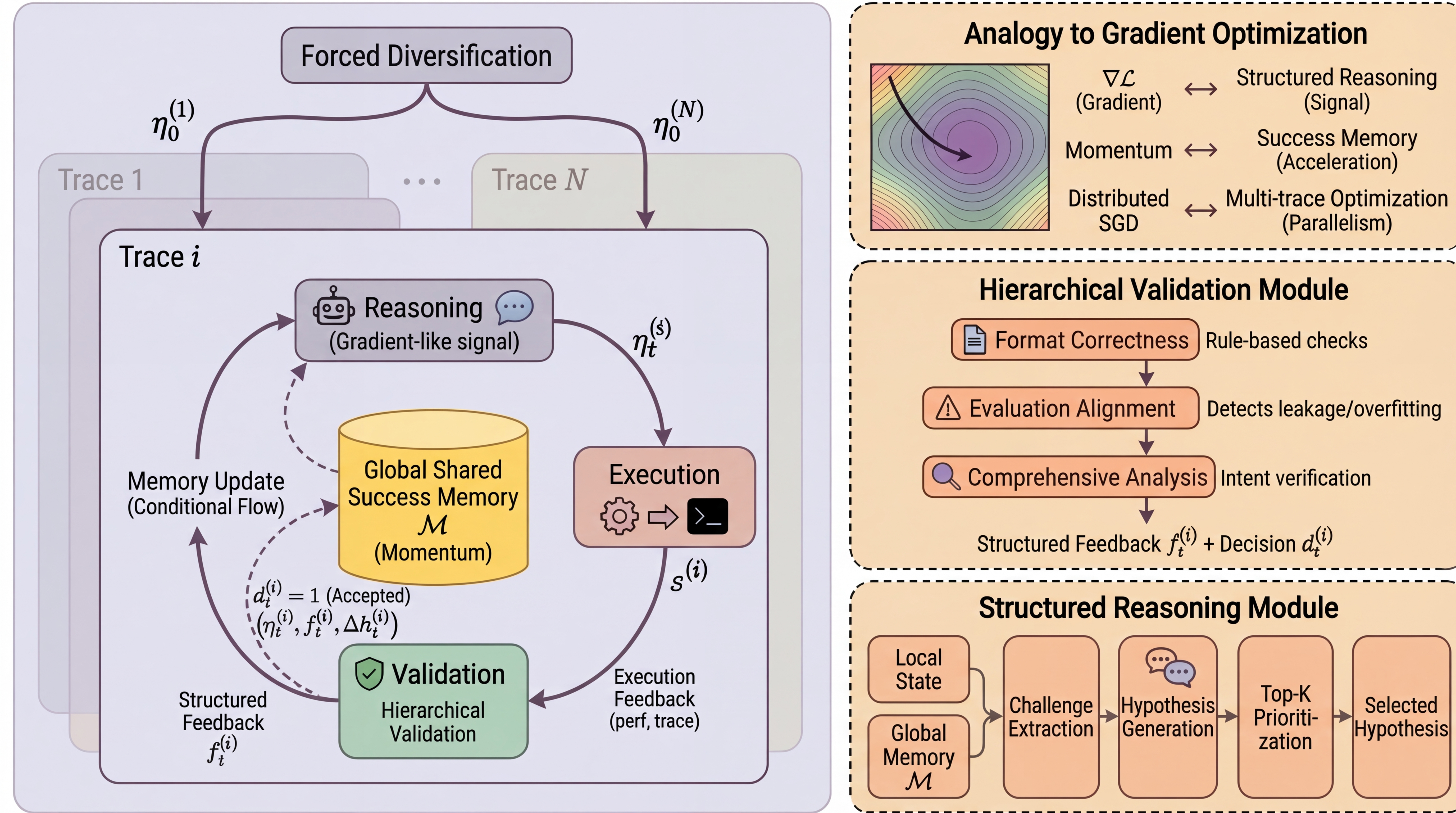}
  \caption{Overview of the \textsc{Gome} framework. Multiple traces optimize in parallel, synchronizing through a global shared success memory $\mathcal{M}$. Each trace iteratively executes solutions, validates improvements, updates shared memory, and reasons over local-global feedback to generate the next hypothesis. Right panels illustrate the gradient-based optimization analogy and key module details.}
  \label{fig:framework}
\end{figure}

Figure~\ref{fig:framework} illustrates the framework. \textsc{Gome} initializes $N$ parallel optimization traces, each beginning with a distinct hypothesis (i.e., an improvement direction for the ML pipeline) through forced diversification (\S\ref{sec:multitrace}). Each trace maintains local best solution $s^{*(i)}$ and experiment history $\mathcal{H}^{(i)}$, while synchronizing via a shared success memory $\mathcal{M}$. Each iteration proceeds through four stages: (1) \textbf{Execution}: run current solution and collect feedback (\S\ref{sec:feedback});  (2) \textbf{Validation}: apply hierarchical checks to form structured feedback $f_t^{(i)}$ (\S\ref{sec:validation});  (3) \textbf{Memory update}: contribute successful hypotheses to $\mathcal{M}$ (Momentum accumulation) (\S\ref{sec:memory});  (4) \textbf{Reasoning}: generate the next hypothesis $\eta^{(i)}_{t+1}$ by combining local feedback with shared memory (Gradient computation)  (\S\ref{sec:reasoning}). This local-global interplay enables collaborative optimization (\S\ref{sec:multitrace}),
with full algorithmic details in Appendix~\ref{sec:appendix_algo}.

\begin{table}[t]
  \centering
  \small
  \begin{tabular}{@{}lll@{}}
    \toprule
    \textbf{Concept} & \textbf{\textsc{Gome} Component} & \textbf{Functional Role} \\
    \midrule
    Gradient $\nabla L$ & Structured Reasoning & Signal: decides how to update \\
    Momentum & Success Memory & Acceleration via proven patterns \\
    Distributed SGD & Multi-trace Optimization & Parallelism with knowledge sharing \\
    \bottomrule
  \end{tabular}
  \caption{Analogy between \textsc{Gome} and gradient-based optimization.}
  \label{tab:mapping}
\end{table}

\subsection{Execution Feedback}
\label{sec:feedback}

Each iteration begins with a trace executing its current solution and collecting local feedback. Executing solution $s^{(i)}_t$ on trace $i$ produces execution feedback:
\begin{equation}
  (\texttt{perf}_t, \texttt{trace}_t) = \texttt{Execute}(s^{(i)}_t, s^{*(i)}),
\end{equation}
where $s^{*(i)}$ denotes the best solution found by trace $i$ so far. The component $\texttt{perf}_t = (h_t, h^{*(i)}) \in \mathbb{R}^2$ contains current score and trace-local best score as scalar metrics; $\texttt{trace}_t$ captures execution logs (stdout/stderr, runtime logs) and code diffs relative to the trace's current best solution $s^{*(i)}$, enabling diagnosis of implementation issues.

\subsection{Hierarchical Validation}
\label{sec:validation}

A critical challenge in LLM-based optimization is distinguishing genuine improvements from deceptive ones. Solutions may achieve high validation scores through shortcuts such as data leakage or metric gaming. Based on execution feedback, hierarchical validation determines acceptance decision $d^{(i)}_t \in \{0, 1\}$ and diagnostic $\texttt{reason}_t$:
\begin{equation}
  (d^{(i)}_t, \texttt{reason}_t) = \texttt{Validate}(s^{(i)}_t, \texttt{perf}_t, \texttt{trace}_t),
\end{equation}
forming the structured feedback $f^{(i)}_t = (\texttt{perf}_t, \texttt{trace}_t, \texttt{reason}_t)$. The validation process enforces three sequential stages: (1) \textbf{Format correctness:} verifies output schema via rule-based checks; (2) \textbf{Evaluation alignment:} detects data leakage and overfitting risks via LLM analysis, e.g., accessing test labels or future information; (3) \textbf{Comprehensive analysis:} verifies whether the hypothesis achieved its intended effect and assesses code quality. The LLM makes the final accept decision $d^{(i)}_t$ based on this reasoning when $h(s^{(i)}_t) \gtrsim h(s^{*(i)})$, with $\texttt{reason}_t$ capturing the analysis from whichever stage determined the outcome, informing subsequent hypothesis generation. Empirically, hierarchical validation detects 66.7\% of deceptive overfitting attempts, compared to 0\% for purely score-driven baselines (Appendix~\ref{app:overfitting_case}).

\subsection{Success Memory}
\label{sec:memory}

Upon validation, the hypothesis and its feedback are committed to the shared success memory $\mathcal{M}$:
\begin{equation}
  \mathcal{M}_{t+1} =
  \begin{cases}
    \mathcal{M}_t \cup \{(\eta^{(i)}_t, f^{(i)}_t, \Delta h^{(i)}_t)\} & \text{if } d^{(i)}_t = 1 \\
    \mathcal{M}_t & \text{otherwise}
  \end{cases}
\end{equation}

where $\eta^{(i)}_t$ is the hypothesis that led to solution $s^{(i)}_t$, $f^{(i)}_t$ is the structured feedback (\S\ref{sec:validation}), and $\Delta h^{(i)}_t = h^{(i)}_t - h^{*(i)}$ records the score difference (may be negative if accepted for code quality merits). Each memory entry contains the complete context: what was tried (hypothesis), what happened (feedback), and how much it helped (score delta). $\mathcal{M}$ is distinct from the raw local history $\mathcal{H}^{(i)}$: in single-trace settings, it serves as a curated subset of verified successes; in multi-trace settings, it naturally extends to a global repository aggregating discoveries across all workers. Functionally analogous to momentum, this global memory biases future updates toward proven directions.

\subsection{Structured Reasoning}
\label{sec:reasoning}

The reasoning module generates the next improvement hypothesis, serving as the gradient signal that directs optimization. It first extracts challenges $\mathcal{C}^{(i)}_t$ from structured feedback $f^{(i)}_t$ and local history $\mathcal{H}^{(i)}$, shifting from exploratory analysis (early iterations) to targeted diagnosis (addressing specific execution failures). For each identified challenge $c$, the module generates a concrete hypothesis $\eta_c$ by conditioning on the trace's complete local state ($s^{*(i)}, f^{(i)}_t, \mathcal{H}^{(i)}$).

Candidates are scored across dimensions $\mathcal{D}$ (impact, alignment, novelty, feasibility, risk-reward)\footnote{Weights: $(0.4, 0.2, 0.2, 0.1, 0.1)$ respectively.}:
\begin{equation}
  \texttt{score}(\eta) = \sum_{d \in \mathcal{D}} w_d \cdot \texttt{score}_d(\eta, \mathcal{M}),
\end{equation}
where the success memory $\mathcal{M}$ (\S\ref{sec:memory}) modulates scoring: hypotheses resembling past successes gain confidence, while similar failures receive penalties. Rather than greedily selecting the top-scored hypothesis, we sample from top-$k$ to maintain exploration diversity (in multi-trace settings, cross-trace mechanisms further guide selection, see \S\ref{sec:multitrace}). The selected hypothesis is finally implemented as solution $s^{(i)}_{t+1}$ via code generation and iterative refinement.

\subsection{Multi-trace Optimization}
\label{sec:multitrace}

Single-trace optimization risks local optima and cannot recover from poor early decisions. \textsc{Gome} employs $N$ parallel traces that synchronize via $\mathcal{M}$, enabling online knowledge sharing where trace $i$'s successful discoveries immediately inform traces $j \neq i$.

\paragraph{Forced Diversification at Initialization.}
To prevent redundant exploration (which renders memory sharing ineffective), we maximize initial search coverage by strictly enforcing diversity. Trace $n$'s starting hypothesis is conditioned on all prior proposals to ensure orthogonality:
\begin{equation}
  \eta_0^{(n)} = \texttt{GenerateHypothesis}\bigl(\mathcal{T}, \{\eta_0^{(j)}\}_{j<n}\bigr),
\end{equation}
forcing the swarm to investigate distinct regions of the solution space from the outset.

\paragraph{Cross-trace Hypothesis Selection.}
In multi-trace settings, each trace constructs a candidate pool from three complementary sources:
\begin{equation}
  \mathcal{H}_{\text{cand}} = \{\eta_c\}_{c \in \mathcal{C}^{(i)}_t} \cup \{\eta^\star\} \cup \texttt{Sample}(\mathcal{M}),
\end{equation}
where $\{\eta_c\}$ are hypotheses generated by the current trace (\S\ref{sec:reasoning}), $\eta^\star$ is the hypothesis from the highest-scoring iteration in $\mathcal{M}$, and $\texttt{Sample}(\mathcal{M})$ retrieves hypotheses via a probabilistic kernel considering embedding similarity and score diffs. An LLM-based selector then produces the final hypothesis:
\begin{equation}
  \eta^{(i)}_{t+1} = \texttt{LLMSelect}(\mathcal{H}_{\text{cand}}, \{f_\eta\}_{\eta \in \mathcal{M}}),
\end{equation}
where $\{f_\eta\}$ denotes the associated feedback stored alongside each hypothesis. The selector has flexibility to select, modify, or generate based on the candidate pool and historical outcomes. Additional details of the probabilistic interaction kernel are provided in Appendix~\ref{sec:appendix_multitrace}.

\subsection{Robust Implementation}
\label{sec:robust}

We employ several mechanisms to ensure reliable optimization: (1)~\textbf{Development-evaluation separation}: initial iterations run on a small data subset with rapid debug loops, where failed code triggers iterative debugging that analyzes error messages and proposes targeted fixes; final validation executes on full data. (2)~\textbf{Multi-seed selection}: for final submission, we rerun top-$k$ candidates with multiple seeds and select the best-performing result, reducing variance from stochastic LLM outputs. (3)~\textbf{Adaptive time management}: an LLM-based module extends time budget based on task complexity and failure patterns, avoiding premature termination on difficult tasks.
\section{Experiments}
\label{Sec:exp}

We evaluate \textsc{Gome} on MLE-Bench under a rigorous closed-world protocol to validate our claim that gradient-based optimization scales more effectively than search as reasoning capability improves. Additional experiments including protocol comparison and cost analysis are provided in Appendix~\ref{app:more_exp}.
Following prior MLE-Bench evaluations, we adopt any-medal rate as the primary metric. We additionally report improvement rate and validation-test correlation (IC) to characterize per-step optimization dynamics.

\subsection{Experiment Setup}

\paragraph{Benchmark.} All experiments are conducted on MLE-Bench~\citep{chan2024mle-bench}, a comprehensive benchmark comprising 75 Kaggle competitions for evaluating AI agents on machine learning engineering tasks, categorized by complexity into low, medium, and high. MLE-Bench-Lite consists of 22 low-complexity tasks for efficient evaluation. We report any-medal rate as the primary indicator.

\paragraph{Implementation.} We evaluate \textsc{Gome} using three frontier LLMs representing a spectrum of reasoning capabilities: DeepSeek-R1~\citep{guo2025deepseek}, o3~\citep{openai_introducing_o3_2025}, and GPT-5~\citep{singh2025openai}, ordered by increasing reasoning strength. For the test environment, we use 12 vCPUs, 220GB RAM, and 1 NVIDIA V100 GPU with a 12-hour time budget. All results are averaged over three runs with different random seeds.

\paragraph{Baselines.} We benchmark against leading closed-world MLE agents on MLE-Bench, including MLAB~\citep{huang2023mlagentbench}, OpenHands~\citep{wang2024openhands}, AIDE~\citep{aide2025}, AIRA~\citep{toledo2025ai}, and ML-Master~\citep{liu2025ml}. For ML-Master, we re-evaluate under identical hardware and time budget to ensure fairness; for other baselines, we use results reported in their respective papers or the official leaderboard.

\subsection{Main Results}

\textbf{\textsc{Gome} establishes a new state-of-the-art on MLE-Bench.} As shown in Table~\ref{tab:main_results}, \textsc{Gome} outperforms all closed-world baselines, achieving a peak any-medal rate of \textbf{35.1\%} with GPT-5. Beyond aggregate performance, \textsc{Gome} excels in solution quality: it achieves a 96.0\% valid submission rate and converts 16.4\% of tasks to Gold medals. Under identical constraints (12h, V100), \textsc{Gome} matches ML-Master on weaker reasoners (DeepSeek-R1: 23.4\% vs. 22.7\%) but pulls ahead significantly with stronger models, widening the gap to \textbf{+11.1 percentage points} with GPT-5 and improving Gold rate. This widening gap is consistent with our hypothesis (validated across 10 models in Section~\ref{sec:scaling}). Furthermore, \textsc{Gome} matches the performance of AIRA despite operating under stricter resource constraints (\textbf{half the time}, weaker GPU), highlighting the efficiency of gradient-based optimization (full cross-protocol comparisons are provided in Table~\ref{tab:protocol_full}).

\textbf{Accurate gradient signals drive rapid convergence.} On MLE-Bench-Lite (Table~\ref{tab:lite}), \textsc{Gome} achieves a \textbf{68.2\%} medal rate, matching the SOTA open-world method (Leeroo) despite lacking external retrieval. This confirms that for tractable tasks, internal diagnostics provide high-fidelity gradients sufficient for rapid convergence~\citep{nesterov2018lectures}. Conversely, the plateau on high-complexity tasks across all methods indicates a ``noisy gradient'' regime where reasoning limits are reached, suggesting that further gains require restoring signal accuracy through stronger models or external knowledge.

Beyond benchmark evaluation, we validate \textsc{Gome} on a live
Kaggle competition, demonstrating its ability to discover non-trivial
feature engineering strategies autonomously (Appendix~\ref{app:kaggle_case}).

\subsection{Ablation Study}

To validate \textsc{Gome}'s design, we ablate its three core components: (1)~\textbf{w/o Structured Reasoning}, replacing diagnostic analysis with simple error-correction prompts; (2)~\textbf{w/o Success Memory}, removing the global repository of proven patterns; and (3)~\textbf{w/o Multi-trace Optimization}, reducing to single-trajectory optimization without cross-trace knowledge sharing.

\begin{table}[ht]
  \centering
  \small
  \setlength{\tabcolsep}{4pt}
  \begin{tabular}{@{}lcc|cccc|ccc@{}}
    \toprule
    & & & \multicolumn{4}{c|}{Medal rate in different complexity (\%)} & \multicolumn{3}{c}{Other evaluation dimensions (\%)} \\
    \cmidrule{4-7} \cmidrule{8-10}
    \textbf{Agent} & \textbf{Time} & \textbf{GPU} & Low & Medium & High & All & Valid & Median+ & Gold \\
    \midrule
    \textbf{MLAB} \\
    gpt-4o-24-08 & 24h & A10$^\dagger$ & 4.2±1.5 & 0.0±0.0 & 0.0±0.0 & 1.3±0.5 & 44.3±2.6 & 1.9±0.7 & 0.8±0.5 \\
    \midrule
    \textbf{OpenHands} \\
    gpt-4o-24-08 & 24h & A10$^\dagger$ & 11.5±3.4 & 2.2±1.3 & 1.9±1.9 & 5.1±1.3 & 52.0±3.3 & 7.1±1.7 & 2.7±1.1 \\
    \midrule
    \textbf{AIDE} \\
    gpt-4o-24-08 & 24h & A10$^\dagger$ & 19.0±1.3 & 3.2±0.5 & 5.6±1.0 & 8.6±0.5 & 54.9±1.0 & 14.4±0.7 & 5.0±0.4 \\
    o1-preview & 24h & A10$^\dagger$ & 34.3±2.4 & 8.8±1.1 & 10.0±1.9 & 16.9±1.1 & 82.8±1.1 & 29.4±1.3 & 9.4±0.8 \\
    \midrule
    \textbf{AIRA} \\
    o3 & 24h & H200 & 55.0±1.5 & \textbf{22.0±1.2} & \underline{21.7±1.1} & 31.6±0.8 & \underline{97.5±0.3} & \textbf{45.5±0.8} & \textbf{17.3±0.4} \\
    \midrule
    \textbf{ML-Master} \\
    DeepSeek-R1 & 12h & V100 & 47.0±6.6 & 9.7±2.3 & 20.0±3.8 & 22.7±0.8 & 91.6±0.4 & 32.9±0.4 & 12.4±0.4 \\
    o3 & 12h & V100 & 42.4±4.0 & 12.3±1.8 & 20.0±0.0 & 22.7±1.3 & 94.2±0.4 & 31.5±2.2 & 13.3±0.8 \\
    GPT-5 & 12h & V100 & 51.5±3.0 & 10.5±2.6 & 17.8±2.2 & 24.0±2.3 & \textbf{98.2±0.4} & 34.7±0.4 & 12.4±1.6 \\
    \midrule
    \textbf{\textsc{Gome} (ours)} \\
    DeepSeek-R1 & 12h & V100 & 50.0±0.0 & 9.7±0.9 & 20.0±0.0 & 23.4±0.4 & 88.9±0.4 & 35.1±0.4 & 12.4±0.9 \\
    o3 & 12h & V100 & \underline{59.1±4.5} & 20.2±0.9 & \textbf{22.2±2.2} & \underline{32.5±1.2} & 94.2±0.4 & 43.1±0.9 & \textbf{17.3±0.8} \\
    GPT-5 & 12h & V100 & \textbf{68.2±2.6} & \underline{21.1±1.5} & \textbf{22.2±2.4} & \textbf{35.1±0.4} & 96.0±0.0 & \underline{45.3±0.0} & \underline{16.4±0.8} \\
    \bottomrule
  \end{tabular}
  \caption{Percentage of achieving any medals across different ML task complexity levels (left) and other evaluation dimensions (right) on MLE-Bench. Reporting results are mean $\pm$ SEM over 3 seeds. Valid, Median+, and Gold indicate the percentage of submissions with valid score, above median score, and gold medal. Best performances are marked in \textbf{bold}; second best are \underline{underlined}. $^\dagger$GPU configuration from official MLE-Bench evaluation setup.}
  \label{tab:main_results}
\end{table}

\begin{table}[H]
  \centering
  \small
  \begin{tabular}{@{}lc@{}}
    \toprule
    \textbf{Agent} & \textbf{Any-Medal (\%)} \\
    \midrule
    KompeteAI$^\dagger$ (Gemini-2.5-flash,6h) & 51.5±1.5 \\
    AIRA (o3,24h) & 55.0±1.5 \\
    InternAgent-MLE$^\dagger$ (DeepSeek-R1,12h) & 62.1±3.0 \\
    FM Agent$^\dagger$ (Gemini-2.5-pro,24h) & 62.1±1.5 \\
    Thesis$^\dagger$ (GPT-5-Codex,24h) & 65.2±1.5 \\
    Leeroo$^\dagger$ (Gemini-3-pro,24h) & \textbf{68.2±2.6} \\
    \textsc{Gome} (GPT-5,12h) & \textbf{68.2±2.6} \\
    \bottomrule
  \end{tabular}
  \caption{Performance comparison on MLE-Bench-Lite. Best performances are marked in \textbf{bold}. $^\dagger$Methods using open-world protocol with external resource access.}
  \label{tab:lite}
\end{table}

\begin{table}[H]
  \centering
  \begin{tabular}{@{}lcccc@{}}
    \toprule
    \textbf{Configuration} & \textbf{Medal} & \textbf{Gold} & \textbf{Impr.} & \textbf{IC} \\
    \midrule
    \textbf{\textsc{Gome} (full)} & \textbf{35.1} & \textbf{16.4} & \textbf{41.1} & \textbf{0.92} \\
    \midrule
    w/o Structured Reasoning & 25.8 & 13.3 & 22.6 & 0.83 \\
    w/o Success Memory & 28.9 & 16.9 & 36.2 & 0.87 \\
    w/o Multi-trace Optimization & 32.4 & 15.1 & 41.3 & 0.88 \\
    \bottomrule
  \end{tabular}
  \caption{Ablation study on \textsc{Gome} components using GPT-5 under 12-hour budget (averaged over 3 runs). Impr.: improvement rate, the percentage of iterations where the proposed solution passes hierarchical validation. IC: Spearman correlation between validation and test improvements.}
  \label{tab:ablation}
\end{table}

As shown in Table~\ref{tab:ablation}, removing any component compromises the system's overall robustness, particularly the primary Any-Medal rate. The \textbf{Structured Reasoning} ablation causes the most severe degradation, with improvement rate dropping from 41.1\% to 22.6\%, indicating that without diagnostic analysis, most optimization steps fail to produce valid improvements. Removing \textbf{Success Memory} shows a different failure mode: while per-iteration success remains decent, the lack of momentum from proven patterns leads to redundant exploration, reducing medal rate by 6.2\%. Finally, \textbf{Multi-trace Optimization} is essential for diversity: its removal degrades final performance despite high local improvement rates, suggesting that cross-trace sharing helps the swarm escape the local optima that trap single-trace agents. All three components contribute meaningfully (Table~\ref{tab:mapping}).

\section{Scaling Analysis}
\label{sec:scaling}

\begin{figure*}[t]
  \centering
  \begin{subfigure}[b]{0.43\linewidth}
    \centering
    \includegraphics[width=\linewidth]{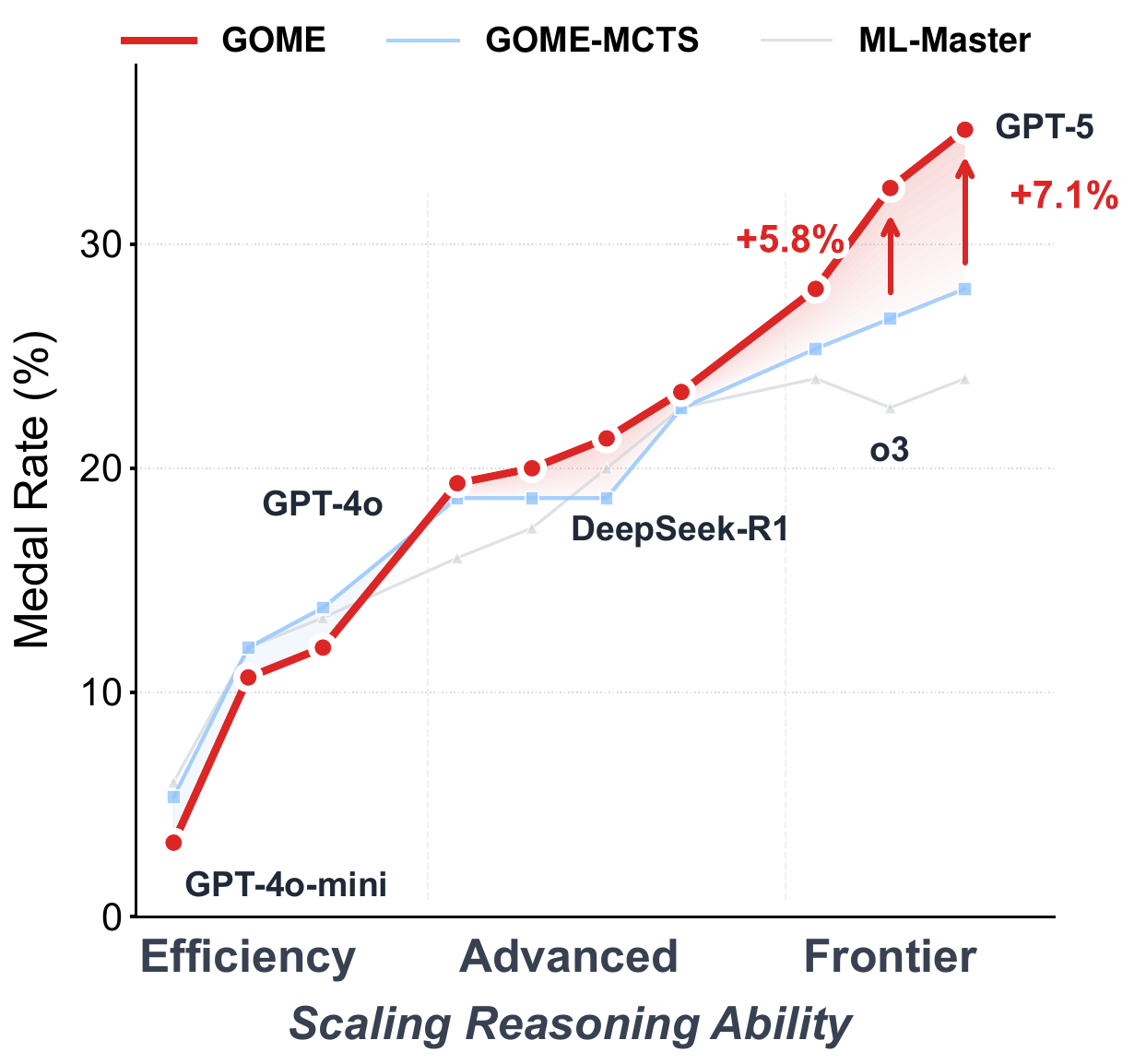}
    \caption{\textbf{Scaling with Model Capability.} As model capability increases from Efficiency to Frontier tiers, \textsc{Gome}'s advantage over search-based baselines widens significantly.}
    \label{fig:scaling_law}
  \end{subfigure}
  \hfill
  \begin{subfigure}[b]{0.52\linewidth}
    \centering
    \includegraphics[width=\linewidth]{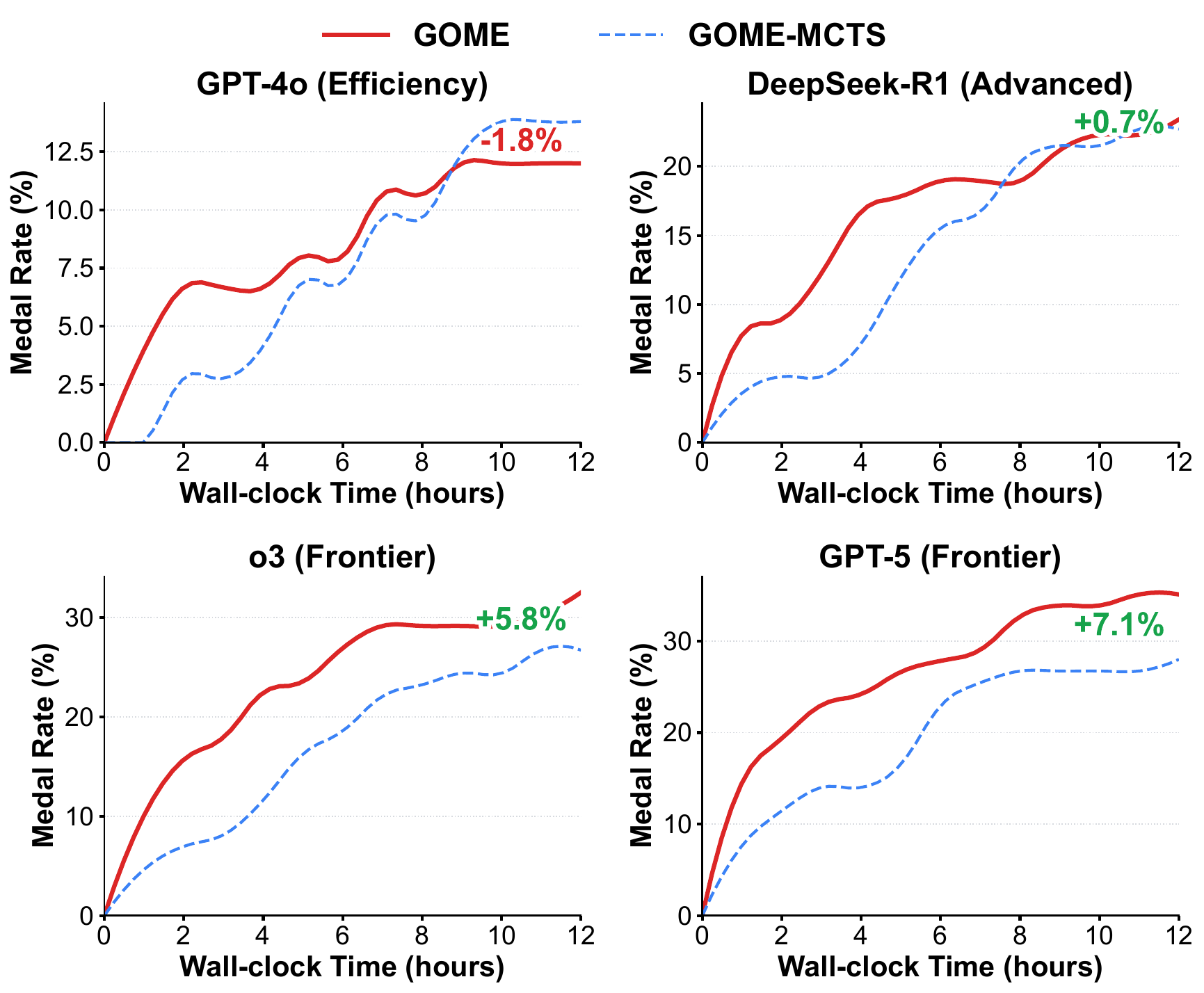}
    \caption{\textbf{Convergence Dynamics.} \textsc{Gome} (solid red) exhibits rapid early convergence, while MCTS (dashed blue) starts slower but catches up on weaker models. On Frontier-tier models, \textsc{Gome} maintains its advantage throughout.}
    \label{fig:dynamics}
  \end{subfigure}
  \caption{\textbf{Scaling analysis.} (a) As reasoning capability improves, gradient-based optimization increasingly outperforms search baselines. (b) On weaker models, rapid early convergence is followed by a plateau due to noisy feedback; stronger reasoning enables sustained improvement throughout the 12h budget.}
  \label{fig:learning-dynamics}
\end{figure*}

To validate our claim that gradient signal quality depends on reasoning capability, we evaluate \textsc{Gome} across three tiers: Efficiency (e.g., GPT-4o-mini), Advanced (e.g., DeepSeek-R1), and Frontier (e.g., o3, GPT-5)\footnote{See Appendix~\ref{app:scaling} for complete tier definitions and per-model results.}. We compare against \textsc{Gome}-MCTS, a controlled variant replacing gradient updates with tree search within identical prompt and infrastructure, isolating the impact of the optimization strategy (implementation details in Appendix~\ref{app:mcts}).

Figure~\ref{fig:scaling_law} reveals a distinct phase transition. On Efficiency-tier models, \textsc{Gome} lags behind: when gradient signals are noisy due to weak reasoning, exhaustive search outperforms directed updates. However, a \textbf{crossover} emerges as capability increases---search gains plateau while \textsc{Gome} accelerates, widening the gap to +5.8\% on o3 and +7.1\% on GPT-5. Figure~\ref{fig:dynamics} traces this divergence temporally. \textsc{Gome} exhibits rapid early convergence across all tiers, but sustainability differs: on weaker models (GPT-4o), unreliable gradients cause premature convergence, allowing MCTS to eventually catch up; on Frontier models, accurate gradient signals enable \textsc{Gome} to maintain its lead throughout, achieving both faster convergence and superior final performance. Complete per-model results and tier-wise gap visualization are provided in Appendix~\ref{app:scaling}.

These results validate our central claim: gradient-based optimization increasingly outperforms tree search as reasoning capability improves. More broadly, they reveal a fundamental divergence in scaling properties: tree search scales with \textit{inference compute} (traversing more nodes), while gradient-based optimization scales with \textit{model capability} (reasoning better). Given the rapid advancement of reasoning-oriented LLMs, this positions gradient-based optimization as the increasingly favorable paradigm for MLE agents.

\section{Conclusion}
We presented \textsc{Gome}, reframing MLE agents through gradient-based optimization. By treating LLM reasoning as gradient signals, success memory as momentum, and multi-trace execution as distributed optimization, \textsc{Gome} establishes a principled correspondence between classical optimization and agentic ML engineering.

Our experiments demonstrate state-of-the-art performance on MLE-Bench, while our scaling analysis reveals a more fundamental finding: tree search scales with inference compute (traversing more nodes), while \textsc{Gome} scales with model capability through increasingly accurate gradient signals. As foundation models advance in reasoning ability, this positions gradient-based optimization as an increasingly favorable paradigm.

This work opens a new design axis for MLE agents: rather than engineering more sophisticated search strategies, future systems may benefit from investing in gradient quality through richer feedback signals and stronger diagnostic reasoning. Future work may also explore hybrid strategies that combine tree search for early-stage exploration with gradient-based refinement as reasoning quality improves within a run. We release our codebase and GPT-5 execution traces to support reproducibility and future research.

\section*{Limitations}

\paragraph{Dependency on Reasoning Capability.}
\textsc{Gome}'s effectiveness is sensitive to the base model's reasoning capability. As shown in our scaling analysis, \textsc{Gome} underperforms tree-search methods on efficiency-tier models, where weaker reasoners may hallucinate diagnostic feedback. This also implies higher inference costs when using frontier models (Table~\ref{tab:cost}). However, this dependency aligns with the trajectory of LLM development: as reasoning-oriented models advance rapidly, the capability threshold for effective gradient-based optimization will become increasingly accessible.

\paragraph{Evaluation Scope.}
We evaluated under a closed-world protocol on MLE-Bench, restricting agents to task-provided materials without external knowledge retrieval. While this ensures rigorous comparison of optimization strategies by controlling for confounding factors such as retrieval quality and knowledge source coverage, it limits generalizability: real-world MLE tasks often benefit from external knowledge, and \textsc{Gome}'s performance in open-world settings remains less thoroughly validated. Additionally, real Kaggle competitions allow participants to access shared notebooks and public baselines, so our closed-world setting may underestimate the performance achievable in practical scenarios.

\paragraph{Lack of Formal Convergence Guarantees.}
Our gradient-based framing establishes functional correspondence rather than mathematical equivalence. Unlike true gradients in continuous optimization, LLM-generated update signals lack formal properties such as convergence guarantees. This means that in rugged landscapes where solutions require fundamentally different paradigms rather than incremental edits, gradient-based guidance may get trapped in local optima. In practice, this is mitigated by the nature of MLE tasks: ML pipelines are typically repairable, making the smoothness assumption reasonable for most practical scenarios. Our multi-trace optimization and success memory further mitigate this risk by enabling shared exploration across trajectories.

%% file: appendix_arxiv.tex
\newpage
\begin{center}

  \LARGE{\textbf{Content of Appendix}}
\end{center}

{
  \hypersetup{linktoc=page}
  \startcontents[sections]
  \printcontents[sections]{l}{1}{\setcounter{tocdepth}{2}}
}

\clearpage

\section*{Appendix}

\section{More Experimental Analysis}
\label{app:more_exp}

\subsection{Extended Scaling Analysis}
\label{app:scaling}

This section provides detailed per-model results supporting the scaling analysis in Section~\ref{sec:scaling}. We categorize models into three capability tiers following the stratification established by the Artificial Analysis Intelligence Index\footnote{https://artificialanalysis.ai/}, SWE-bench, and Aider leaderboard\footnote{https://aider.chat/docs/leaderboards/}~\citep{artificialanalysis2026, jimenez2023swe, aider2025}: \textbf{Efficiency} models are optimized for throughput and cost; \textbf{Advanced} models represent balanced capability-efficiency trade-offs; \textbf{Frontier} models define current state-of-the-art reasoning.

\begin{table}[htbp]
  \centering
  \begin{tabular}{lcccc}
    \toprule
    \textbf{Model} & \textsc{Gome} & \textsc{Gome-MCTS} & \textbf{ML-Master} & \textbf{Runs} \\
    \midrule
    \textbf{Efficiency Tier Models} \\
    GPT-4o-mini & 3.3 & 5.3 & \textbf{6.0} & 2 \\
    GPT-5-nano & 10.7 & \textbf{12.0} & \textbf{12.0} & 2 \\
    GPT-4o & 12.0 & \textbf{13.8} & 13.3 & 3 \\
    \midrule
    \textbf{Advanced Tier Models} \\
    GPT-5-mini & \textbf{19.3} & 18.7 & 16.0 & 2 \\
    Grok-4 & \textbf{20.0} & 18.7 & 17.3 & 2 \\
    Grok-4.1 & \textbf{21.3} & 18.7 & 20.0 & 2 \\
    DeepSeek-R1 & \textbf{23.4} & 22.7 & 22.7 & 3 \\
    \midrule
    \textbf{Frontier Tier Models} \\
    DeepSeek-v3.2 & \textbf{28.0} & 25.3 & 24.0 & 2 \\
    o3 & \textbf{32.5} & 26.7 & 22.7 & 3 \\
    GPT-5 & \textbf{35.1} & 28.0 & 24.0 & 3 \\
    \bottomrule
  \end{tabular}
  \caption{Complete per-model scaling results. All values are any-medal rates (\%) on MLE-Bench. \textbf{Bold} indicates the best performance per model. Runs indicates the number of independent trials.}
  \label{tab:scaling_full}
\end{table}

\begin{figure}[htbp]
  \centering
  \includegraphics[width=0.8\columnwidth]{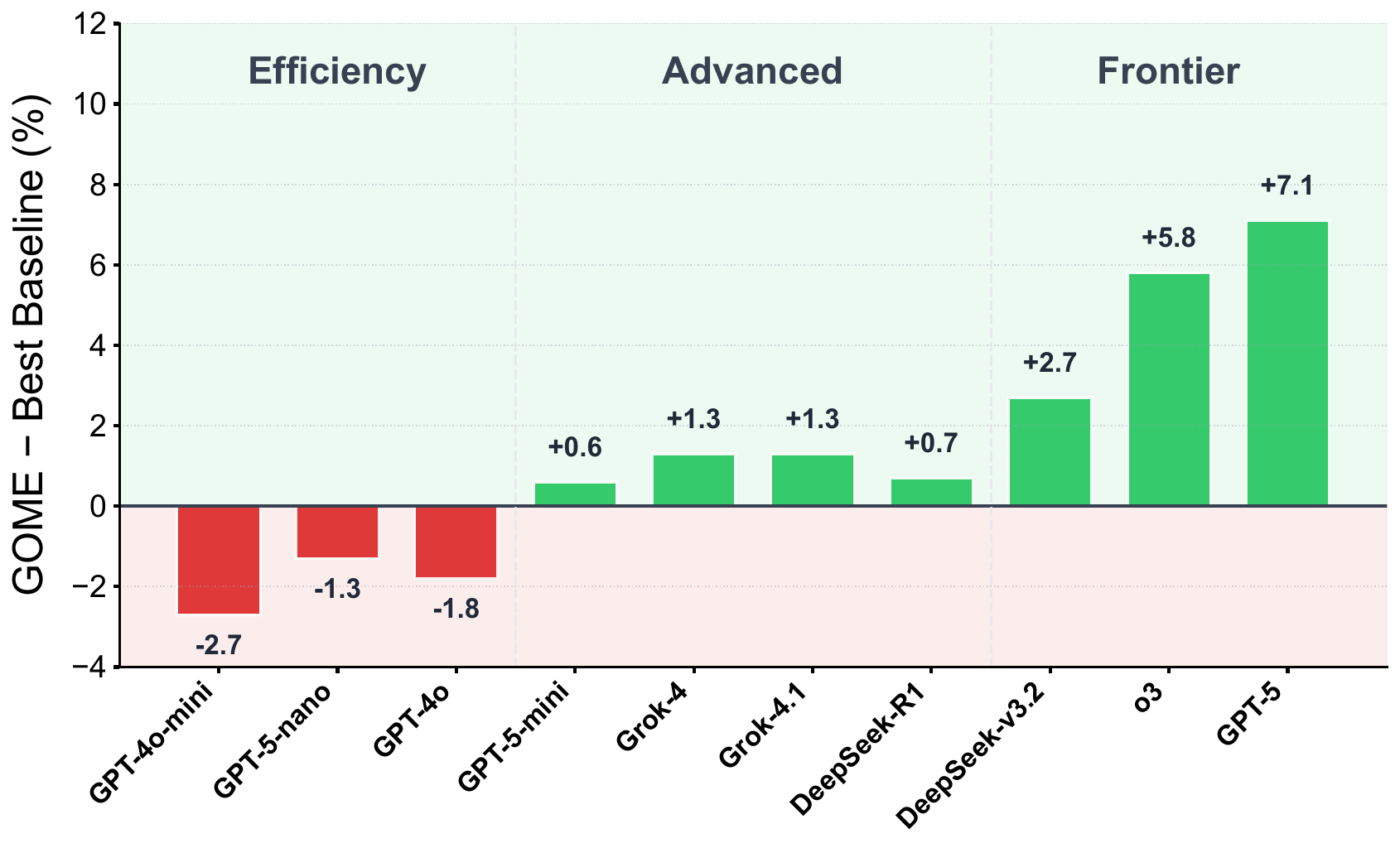}
  \caption{Performance gap between \textsc{Gome} and the best baseline across model capability tiers. Negative values (red) indicate \textsc{Gome} underperforms; positive values (green) indicate \textsc{Gome} outperforms. The crossover occurs at the Advanced tier, with the gap widening progressively in the Frontier tier.}
  \label{fig:scaling_gap}
\end{figure}

Table~\ref{tab:scaling_full} and Figure~\ref{fig:scaling_gap} report the complete results for all ten models evaluated in our scaling analysis. On Efficiency-tier models, \textsc{Gome} underperforms the best baseline consistently, with gaps ranging from $-1.3\%$ to $-2.7\%$. This confirms that weaker models produce unreliable gradient signals. The high sensitivity of gradient-based optimization causes it to be confidently misguided by hallucinated diagnostics, while search-based methods remain robust through exhaustive exploration. The crossover emerges in the Advanced tier, where all four models show positive gaps ranging from $+0.6\%$ to $+1.3\%$. The advantage is modest but consistent, suggesting that mid-capability models begin producing sufficiently accurate gradient signals to offset the exploration benefits of tree search. At the Frontier tier, \textsc{Gome}'s advantage becomes substantial and the gap widens progressively from $+2.7\%$ to $+7.1\%$, demonstrating that stronger reasoners amplify \textsc{Gome}'s relative gains rather than merely maintaining them.

These results validate our central claim: gradient-based optimization exhibits fundamentally different scaling properties from tree search. The performance gap transitions from negative to positive and continues to widen as model capability increases, implying that gradient-based optimization will become increasingly favorable as foundation models advance.

\subsection{Additional Ablations}
\label{app:ablations}

Our main experiments use a 12-hour time budget (half of MLE-Bench's standard 24-hour allocation) for computational efficiency. To examine whether \textsc{Gome} continues to improve with the full time budget, we select o3 and GPT-5 as representative Frontier-tier models and extend to 24 hours.

\begin{table}[htbp]
  \centering
  \begin{tabular}{lccc}
    \toprule
    \textbf{Model} & \textbf{12h} & \textbf{24h} & \textbf{Improvement} \\
    \midrule
    o3 & 32.5 & 36.4 & +3.9 (+12.0\%) \\
    GPT-5 & 35.1 & 40.4 & +5.3 (+15.1\%) \\
    \bottomrule
  \end{tabular}
  \caption{\textsc{Gome} performance under half-time (12h) and full-time (24h) budgets (averaged over 3 runs). All values are any-medal rates (\%) on MLE-Bench. See Table~\ref{tab:protocol_full} for full results.}
  \label{tab:extended_time}
\end{table}

Table~\ref{tab:extended_time} shows that \textsc{Gome} continues to benefit from additional optimization time without plateauing. Both models achieve substantial gains when doubling the time budget, with o3 improving by 12.0\% and GPT-5 by 15.1\% relatively. Notably, the stronger model (GPT-5) exhibits larger improvements, consistent with our scaling claim: higher-quality gradient signals enable more sustained optimization progress. This suggests that gradient-based optimization has not yet saturated its potential on current Frontier models, and further gains may be achievable with increased compute budgets.

\subsection{Impact of External Knowledge Access}
\label{app:protocol}

Table~\ref{tab:main_results} presents results under two evaluation protocols. Methods marked with \ding{51} operate in a closed-world setting, where agents cannot access external knowledge sources during evaluation but may employ carefully designed prompts, multi-stage pipelines, and framework-level optimizations. Methods marked with \ding{55} operate in an open-world setting, where agents can additionally leverage external resources such as web search, documentation retrieval, and curated solution libraries. We include both for comprehensive comparison.

\paragraph{Rationale for Closed-World Evaluation.}
Our main experiments (Section~\ref{Sec:exp}) adopt the closed-world protocol for three reasons. First, it isolates architectural contributions from confounding factors, ensuring that performance differences can be attributed to the optimization framework rather than retrieval quality or knowledge source coverage. Second, it enables fair comparison across methods, as open-world results depend heavily on which external APIs are available, how retrieved content is filtered, and the quality of curated knowledge bases. Third, it directly tests our core claim that improved LLM reasoning translates to better gradient signals, whereas external knowledge access would obscure this relationship by providing an alternative path to performance gains.

\paragraph{Performance Analysis.}
External knowledge access shows varying impact across task complexity. On low-complexity tasks, \textsc{Gome} achieves the highest medal rate across both protocols, suggesting that strong reasoning can fully compensate for the absence of external resources on well-defined problems. The gap widens on medium and high-complexity tasks, where open-world methods benefit from domain-specific knowledge retrieval for specialized competitions. Despite this inherent disadvantage, \textsc{Gome} remains highly competitive, achieving top-three overall performance while operating under strictly closed-world conditions with shorter time budgets and weaker hardware than leading open-world methods.

\begin{table*}[htbp]
  \centering
  \small
  \setlength{\tabcolsep}{6pt}
  \resizebox{0.95\textwidth}{!}{%
    \begin{tabular}{@{}lccc|cccc@{}}
      \toprule
      & & & & \multicolumn{4}{c}{Medal rate in different complexity (\%)} \\
      \cmidrule{5-8}
      \textbf{Agent} & \textbf{w/o Ext. Know.} & \textbf{Time} & \textbf{GPU} & Low & Medium & High & All \\
      \midrule
      \textbf{MLAB} \\
      gpt-4o-24-08 & \ding{51} & 24h & A10$^\ddagger$ & 4.2±1.5 & 0.0±0.0 & 0.0±0.0 & 1.3±0.5 \\
      \midrule
      \textbf{OpenHands} \\
      gpt-4o-24-08 & \ding{51} & 24h & A10$^\ddagger$ & 11.5±3.4 & 2.2±1.3 & 1.9±1.9 & 5.1±1.3 \\
      \midrule
      \textbf{AIDE} \\
      gpt-4o-24-08 & \ding{51} & 24h & A10$^\ddagger$ & 19.0±1.3 & 3.2±0.5 & 5.6±1.0 & 8.6±0.5 \\
      o1-preview & \ding{51} & 24h & A10$^\ddagger$ & 34.3±2.4 & 8.8±1.1 & 10.0±1.9 & 16.9±1.1 \\
      \midrule
      \textbf{AIRA} \\
      o3 & \ding{51} & 24h & H200 & 55.0±1.5 & 22.0±1.2 & 21.7±1.1 & 31.6±0.8 \\
      \midrule
      \textbf{ML-Master} \\
      DeepSeek-R1 & \ding{51} & 12h & V100 & 47.0±6.6 & 9.7±2.3 & 20.0±3.8 & 22.7±0.8 \\
      o3 & \ding{51} & 12h & V100 & 42.4±4.0 & 12.3±1.8 & 20.0±0.0 & 22.7±1.3 \\
      GPT-5 & \ding{51} & 12h & V100 & 51.5±3.0 & 10.5±2.6 & 17.8±2.2 & 24.0±2.3 \\
      \midrule
      \textbf{KompeteAI} \\
      Gemini-2.5-flash & \ding{55} & 6h & A100 & 51.5±1.5 & -- & -- & -- \\
      \midrule
      \textbf{Neo} \\
      Multi Frontier-Tier Models & \ding{55} & 36h & A100 & 48.5±1.5 & 29.8±2.3$^\star$ & 24.4±2.2 & 34.2±0.9 \\
      \midrule
      \textbf{InternAgent-MLE} \\
      DeepSeek-R1 & \ding{55} & 12h & A800 & 62.1±3.0 & 26.3±2.6 & 24.4±2.2 & 36.4±1.2 \\
      \midrule
      \textbf{MLE-STAR} \\
      Gemini-2.5-pro & \ding{55} & 12h & 8*V100 & 63.6±6.0 & -- & -- & -- \\
      \midrule
      \textbf{FM Agent} \\
      Gemini-2.5-pro & \ding{55} & 12h & A800 & 62.1±1.5 & \underline{36.8±1.5} & \textbf{33.3±0.0} & \underline{43.6±0.8} \\
      \midrule
      \textbf{Thesis} \\
      GPT-5-Codex & \ding{55} & 24h & H100 & 65.2±1.5$^\star$ & \textbf{45.6±7.2} & \underline{31.1±2.2} & \textbf{48.4±3.6} \\
      \midrule
      \textbf{\textsc{Gome} (ours)} \\
      DeepSeek-R1 & \ding{51} & 12h & V100 & 50.0±0.0 & 9.7±0.9 & 20.0±0.0 & 23.4±0.4 \\
      o3 & \ding{51} & 12h & V100 & 59.1±4.5 & 20.2±0.9 & 22.2±2.2 & 32.5±1.2 \\
      o3 & \ding{51} & 24h & V100 & 63.6±2.6 & 27.2±2.3 & 24.4±2.2 & 36.4±0.4 \\
      GPT-5 & \ding{51} & 12h & V100 & \underline{68.2±2.6} & 21.1±1.5 & 22.2±2.4 & 35.1±0.4 \\
      GPT-5 & \ding{51} & 24h & V100 & \textbf{71.2±1.5} & 28.1±0.9 & 26.7±0.0$^\star$ & 40.4±0.9$^\star$ \\
      \bottomrule
    \end{tabular}
  }
  \caption{Percentage of achieving any medals across different ML task complexity levels on MLE-Bench. Reporting results are mean $\pm$ SEM over 3 seeds. Best performances are marked in \textbf{bold}; second best are \underline{underlined}; third best are marked with $^\star$. ``--'' indicates results not reported in the original work. $^\ddagger$GPU configuration from official MLE-Bench evaluation setup.}
  \label{tab:protocol_full}
\end{table*}

\subsection{Cost Analysis}
\label{app:cost}

Table~\ref{tab:cost} reports API pricing and average cost per competition for Frontier-tier models under the 12-hour evaluation budget.

\begin{table}[H]
  \centering
  \begin{tabular}{lcccc}
    \toprule
    \textbf{Model} & \textbf{Input} & \textbf{Output} & \textbf{Cost/Task} & \textbf{Medal Rate} \\
    \midrule
    DeepSeek-v3.2 & \$0.28 (\$0.028) & \$0.42 & \$1.89 & 28.0\% \\
    GPT-5 & \$1.25 & \$10.00 & \$20.74 & 35.1\% \\
    o3 & \$2.00 & \$8.00 & \$26.53 & 32.5\% \\
    \bottomrule
  \end{tabular}
  \caption{API pricing (per million tokens) and average cost per competition for Frontier-tier models. Cost/Task is averaged over 3 independent runs. Parenthetical value indicates cache read price.}
  \label{tab:cost}
\end{table}

DeepSeek-v3.2 achieves 28.0\% medal rate at \$1.89 per task, offering the best cost-efficiency among Frontier models. Its pricing is an order of magnitude lower than alternatives, with cache read further reducing input cost to \$0.028/M tokens for repeated context. GPT-5 achieves the highest performance (35.1\%) at \$20.74 per task, where the high output price (\$10/M tokens) dominates the cost. o3 incurs the highest total cost (\$26.53) while underperforming GPT-5, likely due to extended reasoning traces that increase token consumption without proportional performance gains. These results suggest that model selection should balance performance requirements against budget constraints, with DeepSeek-v3.2 as a strong choice for cost-sensitive deployments.

\subsection{Task-type Breakdown}
\label{app:tasktype}

Table~\ref{tab:tasktype} presents detailed score comparison across 9 representative MLE-Bench tasks, grouped by data modality. We evaluate \textsc{Gome} with three frontier-tier backbone models: DeepSeek-R1~\citep{guo2025deepseek}, o3~\citep{openai_introducing_o3_2025}, and GPT-5~\citep{singh2025openai}.

\begin{table*}[htbp]
  \centering
  \small
  \begin{tabular}{llccc}
    \toprule
    \textbf{Task} & \textbf{Metric} & \textbf{DeepSeek-R1} & \textbf{o3} & \textbf{GPT-5} \\
    \midrule
    \textbf{Image Tasks} \\
    aptos2019-blindness-detection & QWK $\uparrow$ & 0.8938 & \textbf{0.9243} & 0.9158 \\
    inaturalist-2019-fgvc6 & Top-1 Error $\downarrow$ & 0.2951 & 0.3201 & \textbf{0.2261} \\
    plant-pathology-2021-fgvc8 & F1 $\uparrow$ & \textbf{0.9100} & 0.8949 & 0.9044 \\
    \midrule
    \textbf{Text Tasks} \\
    detecting-insults-in-social-commentary & AUC $\uparrow$ & 0.9384 & \textbf{0.9452} & 0.9176 \\
    jigsaw-toxic-comment-classification & AUC $\uparrow$ & 0.9848 & 0.9856 & \textbf{0.9865} \\
    spooky-author-identification & Logloss $\downarrow$ & 0.3291 & 0.2880 & \textbf{0.2293} \\
    \midrule
    \textbf{Tabular Tasks} \\
    nomad2018-predict-transparent-conductors & RMSLE $\downarrow$ & 0.0615 & \textbf{0.0593} & 0.0618 \\
    stanford-covid-vaccine & MCRMSE $\downarrow$ & 0.2289 & \textbf{0.2263} & 0.3365 \\
    tabular-playground-series-dec-2021 & Accuracy $\uparrow$ & 0.9615 & 0.9631 & \textbf{0.9632} \\
    \bottomrule
  \end{tabular}
  \caption{Score comparison on 9 MLE-Bench tasks across three data modalities. Best result for each task is highlighted in \textbf{bold}. $\uparrow$: higher is better; $\downarrow$: lower is better.}
  \label{tab:tasktype}
\end{table*}

While GPT-5 achieves significantly higher overall medal rate (35.1\%) compared to o3 (32.5\%) and DeepSeek-R1 (23.4\%), the task-level scores reveal a more nuanced picture. GPT-5 achieves the best score on 4 out of 9 tasks, but o3 also wins 4 tasks, and DeepSeek-R1 achieves the top score on plant-pathology despite having the lowest overall medal rate. This discrepancy arises because medal rate measures the proportion of tasks exceeding competition-specific thresholds, not absolute score superiority. GPT-5's advantage lies in consistently crossing medal thresholds across more tasks, rather than dominating every individual competition. The results suggest that different backbone models bring complementary strengths, and \textsc{Gome}'s framework effectively harnesses these diverse capabilities regardless of the underlying LLM.

\section{Implementation Details}
\label{app:implementation}

\subsection{\textsc{Gome}  Hyperparameters}
\label{app:hyperparameters}

Table~\ref{tab:hyperparameters} summarizes the hyperparameter configuration used in \textsc{Gome} experiments. Parameters are organized according to the framework components described in Section~\ref{sec:method}.

\subsection{ML-Master Reproduction}
\label{app:mlmaster}

We reproduce ML-Master~\citep{liu2025ml} as a representative state-of-the-art tree search baseline. We obtain the official implementation from the authors' public repository\footnote{\url{https://github.com/sjtu-sai-agents/ML-Master}} and adapt it to our evaluation protocol.

\begin{table}[hbtp]
  \centering
  \small
  \begin{tabular}{lp{8cm}r}
    \toprule
    \textbf{Hyperparameter} & \textbf{Description} & \textbf{Default} \\
    \midrule
    \textbf{LLM Configuration} \\
    chat\_model & Base LLM for reasoning & GPT-5 \\
    temperature & LLM decoding temperature & 1.0 \\
    max\_retry & Maximum API retry attempts & 12000 \\
    retry\_wait\_seconds & Wait time between retries (s) & 5 \\
    full\_runtime & Total Running timeout (h) & 12 \\
    \midrule
    \textbf{Success Memory (\S\ref{sec:memory})} \\
    enable\_global\_memory & Enable shared success memory $\mathcal{M}$ across traces & True \\
    memory\_save\_type & Content saved to memory (Full: hypothesis + feedback + score) & Full \\
    \midrule
    \textbf{Structured Reasoning (\S\ref{sec:reasoning})} \\
    llm\_select\_hypothesis & Use LLM-based hypothesis selector & True \\
    simple\_hypothesis & Enable simplified hypothesis format after initialization & True \\
    unique\_hypothesis & Enforce unique hypothesis generation within each trace & True \\
    enable\_cross\_trace\_sharing & Allow hypothesis sampling from other traces via $\mathcal{M}$ & True \\
    \midrule
    \textbf{Multi-trace Optimization (\S\ref{sec:multitrace})} \\
    max\_trace\_num & Number of parallel traces $N$ & 4 \\
    merge\_hours & Hours between trace synchronization & 3 \\
    debugging\_semaphore & Max concurrent debugging operations & 3 \\
    running\_semaphore & Max concurrent running operations & 3 \\
    feedback\_semaphore & Max concurrent feedback operations & 1 \\
    cross\_trace\_diversity & Enforce diversity across traces at initialization & True \\
    \midrule
    \textbf{Robust Implementation (\S\ref{sec:robust})} \\
    coder\_timeout\_multiplier & Upper bound for coder timeout scaling & 4$\times$ \\
    runner\_timeout\_multiplier & Upper bound for runner timeout scaling & 4$\times$ \\
    timeout\_increase\_stage & Initial timeout increase stage & 1 \\
    timeout\_stage\_patience & Patience before timeout escalation & 2 \\
    llm\_decide\_longer\_runtime & Use LLM to decide whether to extend execution time & True \\
    fix\_seed\_and\_split & Fix random seed and data split for reproducibility & True \\
    enable\_multi-seed\_selection & Rerun top candidates with multiple seeds for final submission & True \\
    \bottomrule
  \end{tabular}
  \caption{\textsc{Gome} Hyperparameter Configuration.}
  \label{tab:hyperparameters}
\end{table}

\paragraph{Environment Configuration.}
The original ML-Master experiments were conducted on NVIDIA A100 GPUs. We adapt the implementation to run on NVIDIA V100 GPUs within our standardized Docker environment, ensuring fair comparison across all methods. All experiments use the same MLE-Bench Docker image used for \textsc{Gome}, eliminating potential confounds from environment differences. We apply a 12-hour wall-clock time budget per task, consistent with our closed-world evaluation protocol.

\paragraph{Model Substitution.}
The original ML-Master implementation uses DeepSeek-R1 and GPT-4o as backbone models. To enable scaling analysis across model capability tiers, we modify the inference module to support configurable LLM backends while preserving all other components (tree construction, node selection, and expansion strategies). This allows us to evaluate ML-Master with models ranging from GPT-4o-mini to GPT-5.

\paragraph{Hyperparameters.}
We adopt the default hyperparameters from the original implementation, including search depth, branching factor, and selection criteria. No task-specific tuning is performed to ensure a fair comparison with \textsc{Gome}, which also uses fixed hyperparameters across all tasks.

\paragraph{Reproduction Validation.}
Table~\ref{tab:mlmaster_reproduction} compares our reproduced results with those reported in the original paper. Minor performance differences are attributable to hardware differences (V100 vs.\ A100), Docker environment variations, stochastic variation across runs, and our stricter closed-world protocol that prohibits external knowledge retrieval.

\begin{table}[H]
  \centering
  \begin{tabular}{llcc}
    \toprule
    \textbf{Model} & \textbf{Hardware} & \textbf{Medal Rate} & \textbf{Source} \\
    \midrule
    DeepSeek-R1 & A100 & 29.3 & Official \\
    DeepSeek-R1 & V100 & 22.7 & Our Rerun \\
    \bottomrule
  \end{tabular}
  \caption{ML-Master reproduction validation on DeepSeek-R1. Original results are from \citep{liu2025ml} on A100 GPUs; ours are averaged over 3 independent runs under closed-world protocol on V100 GPUs. All values are any-medal rates (\%) on MLE-Bench with 12-hour budget.}
  \label{tab:mlmaster_reproduction}
\end{table}

\subsection{\textsc{Gome}-MCTS Implementation}
\label{app:mcts}

Our modular framework naturally supports diverse exploration strategies beyond gradient-based optimization. To provide a fair comparison between optimization paradigms within the same framework, we implement Monte Carlo Tree Search (MCTS), a widely-adopted algorithm in existing MLE agents such as ML-Master~\citep{liu2025ml}. This implementation, called \textsc{Gome}-MCTS, serves two purposes: (1) it validates that our framework is a general-purpose platform capable of accommodating different search algorithms, and (2) it isolates the effect of the optimization strategy from other architectural differences when comparing against external baselines.

\paragraph{Architectural Mapping.}
\textsc{Gome}-MCTS reuses \textsc{Gome}'s execution feedback (\S\ref{sec:feedback}), hierarchical validation (\S\ref{sec:validation}), and robust implementation modules (\S\ref{sec:robust}). The key difference lies in how optimization decisions are made: \textsc{Gome}-MCTS replaces structured reasoning (\S\ref{sec:reasoning}) with PUCT-based node selection, and replaces success memory (\S\ref{sec:memory}) with $Q$-value estimates accumulated through tree search. This substitution isolates the comparison to the core optimization mechanism while controlling for other architectural factors.

\paragraph{PUCT Selection.}
We adopt the PUCT (Predictor + UCT) variant with uniform prior $P(n, a) = 1$. The selection score balances exploitation and exploration:
\begin{equation}
  U(n, a) = Q(n, a) + c_{\text{puct}} \cdot \frac{\sqrt{N(n)}}{1 + N(n, a)},
  \label{eq:puct}
\end{equation}
where $Q(n, a)$ represents the average reward of taking action $a$ from node $n$, $N(n, a)$ counts how many times action $a$ has been selected from node $n$, and $N(n)$ is the total visit count for node $n$. The parameter $c_{\text{puct}} \in [0, \infty)$ controls the exploration-exploitation trade-off: $c_{\text{puct}}=0$ yields pure exploitation (greedy selection), while larger values increasingly favor exploration of less-visited actions. We set $c_{\text{puct}} = 1.0$ in all experiments.

\paragraph{Reward Design.}
We experiment with two reward formulations. The binary reward $R_{\text{bin}}$ simply assigns 1 for validated submissions and 0 otherwise:
\begin{equation}
  R_{\text{bin}} =
  \begin{cases}
    1, & \text{if validated}, \\[4pt]
    0, & \text{otherwise}.
  \end{cases}
  \label{eq:reward_binary}
\end{equation}

When a feedback score $v$ is available, score-based reward $R_{\text{score}}$ provides richer signals:
\begin{equation}
  R_{\text{score}} =
  \begin{cases}
    +\tanh(v), & \text{if validated and higher is better}, \\[4pt]
    -\tanh(v), & \text{if validated and lower is better}, \\[4pt]
    -1, & \text{if rejected}.
  \end{cases}
  \label{eq:reward_score}
\end{equation}
The $\tanh$ transformation bounds rewards in $[-1, 1]$ while preserving score ordering. We use $R_{\text{score}}$ in all experiments.

\paragraph{Backpropagation.}
After each rollout, rewards are backpropagated to update $Q$ values along the trajectory using standard averaging:
\begin{equation}
  Q(n, a) \leftarrow \frac{N(n, a) \cdot Q(n, a) + R}{N(n, a) + 1}.
  \label{eq:backprop}
\end{equation}

\paragraph{Tree Expansion.}
Each node expansion generates $k=3$ candidate solutions using the same LLM backbone as \textsc{Gome}. Unlike \textsc{Gome}'s sequential refinement guided by structured reasoning, \textsc{Gome}-MCTS explores multiple branches in parallel and selects based on accumulated $Q$-values. The maximum tree depth is set to 10, with early termination upon achieving a medal-qualifying score.

\section{Algorithm Details}
\label{sec:appendix_algo}

\subsection{Algorithm for Execution Feedback and Hierarchical Validation}

The detailed procedure for generating execution feedback and performing hierarchical validation is presented in Algorithm~\ref{alg:feedback}. This process ensures that the solution $s_t$ is rigorously evaluated against the trace-local best solution $s^*$ through a multi-stage gate mechanism before being accepted.

\begin{algorithm*}[htbp]
  \caption{Execution Feedback and Hierarchical Validation}
  \label{alg:feedback}
  \begin{algorithmic}[1]
    \State \textbf{Input}: current solution $s_t$, trace-local best solution $s^*$, task specification $\mathcal{T}$
    \State \textbf{Output}: structured feedback $f_t$, accept decision $d_t$
    \State
    \State \Comment{\textbf{Stage 1: Execution Feedback} (\S\ref{sec:feedback})}
    \State $h_t \leftarrow \texttt{Evaluate}(s_t, \mathcal{T})$ \Comment{current validation score}
    \State $\texttt{perf}_t \leftarrow (h_t, h^*)$ \Comment{scalar metrics: current and best scores}
    \State $\texttt{trace}_t \leftarrow (\texttt{CodeDiff}(s_t, s^*), \texttt{Logs}(s_t))$ \Comment{code diffs and execution logs}
    \State
    \State \Comment{\textbf{Stage 2: Hierarchical Validation} (\S\ref{sec:validation})}
    \State \Comment{Gate 1: Format correctness (rule-based)}
    \If{$\lnot\, \texttt{FormatValid}(s_t, \mathcal{T})$}
    \State $\texttt{reason}_t \leftarrow \texttt{DiagnoseFormat}(s_t, \mathcal{T})$
    \State \Return $((\texttt{perf}_t, \texttt{trace}_t, \texttt{reason}_t),\, 0)$
    \EndIf
    \State \Comment{Gate 2: Evaluation alignment (LLM-based)}
    \State $\texttt{leakage} \leftarrow \texttt{DetectLeakage}(s_t, \mathcal{T})$ \Comment{test label access, future info, etc.}
    \If{$\texttt{leakage} \neq \emptyset$}
    \State $\texttt{reason}_t \leftarrow \texttt{DiagnoseAlignment}(s_t, \texttt{leakage})$
    \State \Return $((\texttt{perf}_t, \texttt{trace}_t, \texttt{reason}_t),\, 0)$
    \EndIf
    \State \Comment{Gate 3: Comprehensive analysis (when $h_t \gtrsim h^*$)}
    \State $\texttt{reason}_t.\texttt{hyp\_verify} \leftarrow \texttt{VerifyHypothesis}(s_t, s^*, \texttt{trace}_t)$ \Comment{intended effect achieved?}
    \If{$h_t \gtrsim h^*$}
    \State $\texttt{reason}_t.\texttt{code\_quality} \leftarrow \texttt{AnalyzeCode}(s_t, s^*)$ \Comment{best practices, efficiency}
    \EndIf
    \State
    \State \Comment{\textbf{Stage 3: Accept Decision} (LLM-based)}
    \State $f_t \leftarrow (\texttt{perf}_t, \texttt{trace}_t, \texttt{reason}_t)$
    \State $d_t \leftarrow \texttt{LLMJudge}(f_t)$ \Comment{accept based on full structured feedback}
    \State \Return $(f_t,\, d_t)$
  \end{algorithmic}
\end{algorithm*}

\subsection{Algorithm for Structured Reasoning }

The structured reasoning process, which guides the generation of the next hypothesis $\eta_{t+1}$, is outlined in Algorithm~\ref{alg:reasoning}. This algorithm utilizes an adaptive weighting mechanism and leverages the global success memory $\mathcal{M}$ to prioritize hypotheses based on historical success and scenario analysis.

\label{sec:algo_grad_direction}
\begin{algorithm*}[htbp]
  \caption{Structured Reasoning}
  \label{alg:reasoning}
  \begin{algorithmic}[1]
    \State \textbf{Input}: feedback $f_t$, local history $\mathcal{H}$, global success memory $\mathcal{M}$, best-so-far $s^*$, task $\mathcal{T}$
    \State \textbf{Output}: selected hypothesis $\eta_{t+1}$, updated solution $s_{t+1}$
    \State
    \State \Comment{\textbf{Stage 1: Adaptive Weighting}}
    \State $n_{\text{succ}} \leftarrow |\{e \in \mathcal{H} : d_e = 1\}|$
    \State $n_{\text{fail}} \leftarrow |\mathcal{H}| - n_{\text{succ}}$
    \State $\lambda \leftarrow \max(0, 3 - \lfloor (3 n_{\text{succ}} + 2 n_{\text{fail}}) / 8 \rfloor)$ \Comment{decays with experience}
    \State
    \State \Comment{\textbf{Stage 2: Challenge Extraction}}
    \State $\mathcal{C}_t \leftarrow \emptyset$
    \If{$\lambda > 0$} \Comment{early: scenario-driven}
    \State $\mathcal{C}_t \leftarrow \mathcal{C}_t \cup \texttt{AnalyzeScenario}(s^*, \mathcal{T})$ \Comment{SOTA alignment, gap identification}
    \EndIf
    \If{$\lambda < 3$} \Comment{later: feedback-driven}
    \State $\mathcal{C}_t \leftarrow \mathcal{C}_t \cup \texttt{AnalyzeHistory}(f_t, \mathcal{H})$ \Comment{explicit issues, persistent errors}
    \EndIf
    \State
    \State \Comment{\textbf{Stage 3: Hypothesis Generation}}
    \For{$c \in \mathcal{C}_t$}
    \State $\eta_c \leftarrow \texttt{Hypothesize}(c, s^*, f_t, \mathcal{T})$ \Comment{target component + reasoning}
    \EndFor
    \State
    \State \Comment{\textbf{Stage 4: Prioritization}}
    \State $\mathcal{D} \leftarrow \{\texttt{impact}, \texttt{alignment}, \texttt{novelty}, \texttt{feasibility}, \texttt{risk}\}$
    \State $\mathbf{w} \leftarrow (0.4, 0.2, 0.2, 0.1, 0.1)$
    \For{$(c, \eta) \in \mathcal{C}_t$}
    \State $\texttt{score}(\eta) \leftarrow \sum_{d \in \mathcal{D}} w_d \cdot \texttt{score}_d(\eta, \mathcal{M})$
    \EndFor
    \State
    \State \Comment{\textbf{Stage 5: Hypothesis Selection}}
    \If{single-trace}
    \State $\mathcal{H}_{\text{cand}} \leftarrow \texttt{TopK}(\{(c, \eta_c)\}, k)$
    \State $\eta_{t+1} \leftarrow \texttt{Sample}(\mathcal{H}_{\text{cand}})$ \Comment{stochastic selection}
    \Else \Comment{multi-trace (\S\ref{sec:multitrace})}
    \State $\eta^\star \leftarrow \arg\max_{\eta \in \mathcal{M}} \Delta h_\eta$ \Comment{best from shared memory}
    \State $\mathcal{H}_{\text{sim}} \leftarrow \texttt{SimilaritySample}(\mathcal{M}, f_t)$ \Comment{embedding-based retrieval}
    \State $\mathcal{H}_{\text{cand}} \leftarrow \{\eta_c\}_{c \in \mathcal{C}_t} \cup \{\eta^\star\} \cup \mathcal{H}_{\text{sim}}$
    \State $\eta_{t+1} \leftarrow \texttt{LLMSelect}(\mathcal{H}_{\text{cand}}, \{f_\eta\}_{\eta \in \mathcal{M}})$ \Comment{select / modify / generate}
    \EndIf
    \State
    \State \Comment{\textbf{Stage 6: Solution Update}}
    \State $\tau \leftarrow \texttt{Sketch}(\eta_{t+1}, s^*)$ \Comment{code modification plan}
    \State $s_{t+1} \leftarrow \texttt{Implement}(\tau, s^*)$ \Comment{iterative refinement until executable}
    \State \Return $(\eta_{t+1}, s_{t+1})$
  \end{algorithmic}
\end{algorithm*}

\newpage
\subsection{Multi-trace Optimization Details}
\label{sec:appendix_multitrace}

This section provides implementation details for the multi-trace optimization described in \S\ref{sec:multitrace}.

\subsubsection{Candidate Hypothesis Construction}

As described in \S\ref{sec:multitrace}, each trace constructs a candidate pool from three complementary sources:
\begin{equation}
  \mathcal{H}_{\text{cand}} = \{\eta_c\}_{c \in \mathcal{C}^{(i)}_t} \cup \{\eta^\star\} \cup \mathcal{H}_{\text{sim}},
\end{equation}
where:
\begin{itemize}
  \item $\{\eta_c\}_{c \in \mathcal{C}^{(i)}_t}$: Hypotheses generated by the current trace based on local challenges (\S\ref{sec:reasoning})
  \item $\eta^\star$: The hypothesis from the highest-scoring iteration in the shared success memory $\mathcal{M}$
  \item $\mathcal{H}_{\text{sim}}$: Hypotheses sampled via the probabilistic interaction kernel
\end{itemize}

\paragraph{Probabilistic Interaction Kernel.}
Inspired by statistical physics~\citep{isihara2013statistical}, we introduce a probabilistic interaction mechanism that enables controlled information exchange between traces. For each hypothesis $\eta_c$ generated by the current trace, we compute an interaction potential with all hypotheses $\eta_j$ in the shared memory $\mathcal{M}$:
\begin{equation}
  U_{cj} = \alpha \cdot S_{cj} \cdot e^{-\gamma L} + \beta \cdot \tanh(\Delta_{cj})\in [-2, 2],
\end{equation}
where $U_{cj}$ is the interaction potential between hypothesis $\eta_c$ and historical hypothesis $\eta_j$. The parameters $\alpha$ and $\beta$ are weights controlling the relative importance of embedding similarity $S_{cj}$ (cosine similarity between embeddings of $\eta_c$ and $\eta_j$) and score difference $\Delta_{cj}$. The parameter $\gamma$ is a decay factor based on the iteration count $L$.

The score difference $\Delta_{cj}$ is defined as:
\begin{equation}
  \Delta_{cj} =
  \begin{cases}
    h_j - h^{*(i)}, & \text{if higher score is better} \\
    h^{*(i)} - h_j, & \text{if lower score is better}
  \end{cases}
\end{equation}
where $h^{*(i)}$ is the trace-local best score and $h_j$ is the score associated with hypothesis $\eta_j$. The sampling distribution is computed via softmax normalization:
\begin{equation}
  p_{cj} = \frac{\exp(U_{cj})}{\sum_{k} \exp(U_{ck})}, \quad \eta_{\text{sim}} \sim \text{Categorical}(p_{cj}).
\end{equation}

This interaction potential integrates both semantic information from hypothesis text and empirical information from scores. The decay factor $e^{-\gamma L}$ applied to the similarity term reflects that the optimization trajectory is not a Markov process: the generation of later hypotheses depends on multiple previous steps. Therefore, in later stages of exploration, the weight of this component decays rapidly, so that score information plays a more dominant role, biasing selection toward empirically proven strategies.

\subsubsection{Adaptive Hypothesis Selection}

After constructing the candidate pool, an LLM-based selector produces the final hypothesis. The selector is not constrained to choose from the candidates; instead, it can take one of three actions:
\begin{itemize}
  \item \textbf{Select}: Choose the most promising hypothesis from $\mathcal{H}_{\text{cand}}$
  \item \textbf{Modify}: Revise an existing candidate to improve it (e.g., adjust hyperparameters, refine approach)
  \item \textbf{Generate}: Create a new hypothesis by synthesizing insights from multiple candidates
\end{itemize}
This design aims to reduce hallucinations and stabilize outcomes across different traces, as the provided candidates serve as grounded references rather than strict constraints.

Algorithm~\ref{alg:hyp_select} presents the cross-trace hypothesis selection process.

\begin{algorithm*}[htbp]
  \caption{Cross-trace Hypothesis Selection}
  \label{alg:hyp_select}
  \begin{algorithmic}[1]
    \State \textbf{Input}: local hypotheses $\{\eta_c\}_{c \in \mathcal{C}^{(i)}_t}$, success memory $\mathcal{M}$, trace-local best score $h^{*(i)}$
    \State \textbf{Output}: selected hypothesis $\eta_{t+1}^{(i)}$
    \State
    \State \Comment{\textbf{Step 1: Construct Candidate Pool} (\S\ref{sec:multitrace})}
    \State $\eta^\star \leftarrow \arg\max_{\eta \in \mathcal{M}} \Delta h_\eta$ \Comment{best from shared memory}
    \State $\mathcal{H}_{\text{sim}} \leftarrow \texttt{SampleByKernel}(\mathcal{M}, \{\eta_c\})$ \Comment{probabilistic interaction}
    \State $\mathcal{H}_{\text{cand}} \leftarrow \{\eta_c\}_{c \in \mathcal{C}^{(i)}_t} \cup \{\eta^\star\} \cup \mathcal{H}_{\text{sim}}$
    \State
    \State \Comment{\textbf{Step 2: LLM-based Selection}}
    \State $\eta_{t+1}^{(i)} \leftarrow \texttt{LLMSelect}(\mathcal{H}_{\text{cand}}, \{f_\eta\}_{\eta \in \mathcal{M}})$
    \State \Comment{Selector can: \textbf{Select} from candidates, \textbf{Modify} existing, or \textbf{Generate} new}
    \State \Return $\eta_{t+1}^{(i)}$
  \end{algorithmic}
\end{algorithm*}

This adaptive selection mechanism, combined with the probabilistic interaction kernel, enables efficient cross-trace learning without sacrificing exploration diversity. When a trace falls behind the global best, it can leverage successful strategies from other traces via $\eta^\star$ and $\mathcal{H}_{\text{sim}}$; when leading, it continues local exploration while still benefiting from the shared memory. This design ensures that the system leverages collective discoveries from all parallel traces while maintaining independent optimization trajectories.

\subsection{Algorithm for \textsc{Gome} Main Loop}
Algorithm~\ref{alg:gome_main} orchestrates the overall optimization process. It begins with ``Initialization with Forced Diversification'', where starting hypotheses are sequentially conditioned on prior proposals to maximize the initial coverage of the solution space.

The core ``Parallel Optimization'' then coordinates multiple concurrent traces. In each iteration, the system: (1) invokes the feedback mechanism (Algorithm~\ref{alg:feedback}) to evaluate the current state; (2) updates the Global Success Memory $\mathcal{M}$ upon acceptance, allowing successful strategies to be shared across traces; and (3) calls the structured reasoning module (Algorithm~\ref{alg:reasoning}) to generate the next solution. The process concludes with Final Selection, which employs multi-seed evaluation on top candidates to mitigate variance and ensure the robustness of the final output $s^*$.

\begin{algorithm*}[htbp]
  \caption{\textsc{Gome}: Gradient-based Optimization for Machine Learning Engineering}
  \label{alg:gome_main}
  \small
  \begin{algorithmic}[1]
    \State \textbf{Input}: task specification $\mathcal{T}$, dataset $\mathcal{D}$, number of traces $N$, time budget $B$
    \State \textbf{Output}: best solution $s^*$
    \State
    \State \Comment{\textbf{Phase 1: Initialization with Forced Diversification} (\S\ref{sec:multitrace})}
    \State $\mathcal{M} \leftarrow \emptyset$ \Comment{global success memory}
    \For{$n = 1$ to $N$}
    \State $\mathcal{H}^{(n)} \leftarrow \emptyset$ \Comment{local history for trace $n$}
    \State $\eta_0^{(n)} \leftarrow \texttt{GenerateInitialHypothesis}(\mathcal{T}, \{\eta_0^{(j)}\}_{j<n})$ \Comment{condition on earlier proposals}
    \State $s_0^{(n)} \leftarrow \texttt{Implement}(\eta_0^{(n)}, \mathcal{T})$ \Comment{initial solution for trace $n$}
    \State $s^{*(n)} \leftarrow s_0^{(n)}$ \Comment{trace-local best}
    \EndFor
    \State
    \State \Comment{\textbf{Phase 2: Parallel Optimization}}
    \While{$\texttt{RemainingBudget}(B) > 0$}
    \For{each trace $n \in \{1, \ldots, N\}$ \textbf{in parallel}}
    \State
    \State \Comment{Step 1: Execute and Validate (Algorithm~\ref{alg:feedback})}
    \State $(f_t^{(n)}, d_t^{(n)}) \leftarrow \texttt{FeedbackAndValidation}(s_t^{(n)}, s^{*(n)}, \mathcal{T})$
    \State
    \State \Comment{Step 2: Update Success Memory (\S\ref{sec:memory})}
    \If{$d_t^{(n)} = 1$} \Comment{accepted}
    \State $\Delta h_t^{(n)} \leftarrow h_t^{(n)} - h^{*(n)}$
    \State $\mathcal{M} \leftarrow \mathcal{M} \cup \{(\eta_t^{(n)}, f_t^{(n)}, \Delta h_t^{(n)})\}$
    \If{$h_t^{(n)} > h^{*(n)}$}
    \State $s^{*(n)} \leftarrow s_t^{(n)}$ \Comment{update trace-local best}
    \EndIf
    \EndIf
    \State $\mathcal{H}^{(n)} \leftarrow \mathcal{H}^{(n)} \cup \{(s_t^{(n)}, f_t^{(n)}, d_t^{(n)})\}$
    \State
    \State \Comment{Step 3: Structured Reasoning (Algorithm~\ref{alg:reasoning})}
    \State $(\eta_{t+1}^{(n)}, s_{t+1}^{(n)}) \leftarrow \texttt{StructuredReasoning}(f_t^{(n)}, \mathcal{H}^{(n)}, \mathcal{M}, s^{*(n)}, \mathcal{T})$
    \EndFor
    \State
    \State \Comment{Adaptive Time Management (\S\ref{sec:robust})}
    \State $B \leftarrow \texttt{AdjustBudget}(B, \mathcal{M}, \{f_t^{(n)}\}_{n=1}^N)$
    \EndWhile
    \State
    \State \Comment{\textbf{Phase 3: Final Selection} (\S\ref{sec:robust})}
    \State $\mathcal{S}_{\text{top}} \leftarrow \texttt{TopK}(\{s^{*(n)}\}_{n=1}^N, k)$
    \For{$s \in \mathcal{S}_{\text{top}}$}
    \State $\bar{h}(s) \leftarrow \texttt{MultiSeedEval}(s, \mathcal{T})$ \Comment{reduce variance}
    \EndFor
    \State $s^* \leftarrow \arg\max_{s \in \mathcal{S}_{\text{top}}} \bar{h}(s)$
    \State \Return $s^*$
  \end{algorithmic}
\end{algorithm*}

\section{Qualitative Analysis and Case Studies}
\label{sec:appendix_cases}
We provide detailed case studies to illustrate \textsc{Gome}'s reasoning capabilities, contrasting it with scalar-driven baselines and demonstrating its performance in real-world scenarios.

\subsection{Preventing Overfitting via Hierarchical Validation}
\label{app:overfitting_case}

A critical limitation of score-centric feedback utilization is the reliance on validation scores as the sole proxy for solution quality. This often leads to overfitting or the acceptance of technically flawed solutions that happen to score well on a specific split.

We analyze \textsc{Gome}'s overfitting detection capability on the Stanford COVID Vaccine task from MLE-Bench (High tier), where the goal is to predict RNA degradation rates (MCRMSE, lower is better). Across 90 optimization iterations, we identified 9 cases where validation improved but test performance degraded, which is the hallmark of overfitting.

\paragraph{Detection Performance.}
Table~\ref{tab:overfit_stats} summarizes \textsc{Gome}'s overfitting detection results. The hierarchical validation correctly rejected 6 out of 9 overfitting cases (66.7\% detection rate), preventing potentially harmful updates from being committed.

\begin{table}[htbp]
  \centering
  \small
  \begin{tabular}{lc}
    \toprule
    \textbf{Metric} & \textbf{Value} \\
    \midrule
    Total optimization iterations & 90 \\
    Overfitting cases identified & 9 \\
    Correctly rejected & 6 (66.7\%) \\
    Incorrectly accepted & 3 (33.3\%) \\
    \bottomrule
  \end{tabular}
  \caption{Overfitting detection performance on Stanford COVID Vaccine.}
  \label{tab:overfit_stats}
\end{table}

\paragraph{Case Study: Loop 32.}
Figure~\ref{fig:case_overfitting} presents a representative case where \textsc{Gome} correctly rejected a deceptive improvement. The agent proposed a loss reweighting strategy that showed dramatic validation improvement ($-57.6\%$) but would have caused catastrophic test degradation ($+137.5\%$).

\begin{figure*}[htbp]
  \centering
  \begin{tcolorbox}[title=Case: Overfitting Detection on Stanford COVID Vaccine (Node 32), colback=white, colframe=gray!50!black]
    \small
    \textbf{Task}: Predict RNA degradation rates (MCRMSE, lower is better) \\[4pt]
    \textbf{Hypothesis $\eta_t$}: Align training objective with evaluation by redefining loss weights: for reactivity use clipped inverse-variance weights $w=\mathrm{clip}(1/(\mathrm{err}^2+10^{-6}), 0.1, 10)$; for deg\_Mg\_pH10 and deg\_Mg\_50C use constant $w=1.0$. Replace hard SN\_filter gating with soft factor (1.0 for SN=1, 0.5 for SN=0). Per-batch renormalize so each scored channel has equal total weight mass. \\[4pt]
    \textbf{Metrics}:
    \begin{itemize}[nosep]
      \item Previous (Parent 27): $h^{*(i)}_{\mathrm{val}} = 0.6948$, $h^{*(i)}_{\mathrm{test}} = 0.2688$
      \item Current (Node 32): $h_{t,\mathrm{val}} = 0.2945$ (\textcolor{green!50!black}{$\downarrow 57.6\%$ better}), $h_{t,\mathrm{test}} = 0.6382$ (\textcolor{red}{$\uparrow 137.5\%$ worse})
    \end{itemize}
    \vspace{4pt}
    \hrule
    \vspace{4pt}
    \textbf{Code Changes}: Kept PairAwareBiGRU architecture, features, split, inference identical to SOTA. Replaced training weights with channel-specific weighting; introduced per-batch channel renormalization; left validation metric as unweighted MCRMSE. \\[4pt]
    \textbf{Observations from $f_t$}: Current MCRMSE is 0.2945 vs.\ SOTA 0.2559; performance degraded by $+0.0385$. EDA indicates weak target--error association for reactivity (Pearson $\approx -0.042$, Spearman $\approx 0.280$), suggesting inverse-variance weighting may not benefit learning. \\[4pt]
    \textbf{Hypothesis Evaluation}: \textcolor{red}{\textbf{REFUTED}}. The redesigned loss weighting and per-batch renormalization worsened validation MCRMSE. Likely causes: weak correlation between reactivity and its error; soft SN scaling introducing noisier samples; renormalization potentially overemphasizing harder channels. \\[4pt]
    \textbf{Decision}: $d_t = 0$ \textcolor{red}{\textbf{(REJECT)}} --- Despite validation appearing to improve, the structured analysis identified that the ``improvement'' stemmed from metric misalignment rather than genuine generalization gains.
  \end{tcolorbox}
  \caption{Overfitting detection on Stanford COVID Vaccine. The validation score appeared to improve by 57.6\%, but \textsc{Gome}'s hierarchical validation analyzed the code changes and training dynamics to correctly reject the solution, avoiding a catastrophic 137.5\% test regression.}
  \label{fig:case_overfitting}
\end{figure*}

\paragraph{Detailed Overfitting Cases.}
Table~\ref{tab:overfit_cases} lists all identified overfitting cases and \textsc{Gome}'s decisions. The three incorrectly accepted cases (Nodes 40, 49, 80) all exhibited marginal validation improvements ($<6\%$) with small test regressions ($<6.3\%$), making them harder to distinguish from noise. In contrast, the correctly rejected cases often showed larger validation-test divergence patterns.

\begin{table}[htbp]
  \centering
  \small
  \begin{tabular}{cccccc}
    \toprule
    \textbf{Loop} & \textbf{Component} & $\boldsymbol{\Delta h_{\mathrm{val}}}$ & $\boldsymbol{\Delta h_{\mathrm{test}}}$ & \textbf{Decision} & \textbf{Correct} \\
    \midrule
    23 & Model & $-63.8\%$ & $+0.3\%$ & Reject & \ding{51} \\
    26 & FeatureEng & $-5.1\%$ & $+1.4\%$ & Reject & \ding{51} \\
    32 & Model & $-57.6\%$ & $+137.5\%$ & Reject & \ding{51} \\
    40 & Model & $-0.2\%$ & $+6.3\%$ & Accept & \ding{55} \\
    47 & Ensemble & $-11.9\%$ & $+0.4\%$ & Reject & \ding{51} \\
    49 & Model & $-5.8\%$ & $+1.2\%$ & Accept & \ding{55} \\
    57 & Model & $-69.8\%$ & $+0.3\%$ & Reject & \ding{51} \\
    77 & FeatureEng & $-0.1\%$ & $+0.03\%$ & Reject & \ding{51} \\
    80 & Model & $-0.1\%$ & $+0.03\%$ & Accept & \ding{55} \\
    \bottomrule
  \end{tabular}
  \caption{All overfitting cases in Stanford COVID Vaccine. $\Delta h_{\mathrm{val}}$: validation change (negative is improvement). $\Delta h_{\mathrm{test}}$: test change (positive is degradation).}
  \label{tab:overfit_cases}
\end{table}

\paragraph{Analysis.}
This case demonstrates why methods that use feedback primarily for scalar ranking are fundamentally limited: they cannot distinguish genuine improvements from deceptive ones. A score-driven agent would have accepted Loop 32's solution based on the 57.6\% validation ``improvement,'' leading to catastrophic test degradation. \textsc{Gome}'s structured feedback, which analyzes code changes, training dynamics, and feature correlations, provides the semantic information necessary to detect such patterns. The 66.7\% detection rate, while not perfect, represents a significant advantage over purely score-driven approaches that would accept all 9 overfitting cases (0\% detection).

\subsection{Cross-Task Analysis of Deceptive Improvements}
\label{app:deceptive_cross_task}

Our earlier analysis focused on a single representative task. This section extends the analysis across multiple MLE-Bench tasks to identify systematic patterns in deceptive improvements---iterations where validation metrics improve but test performance degrades.

\paragraph{Task-Type Vulnerability.}
Examining GPT-5 optimization traces across the full MLE-Bench, we find that deceptive improvements are not uniformly distributed but concentrate in two task categories:

\begin{enumerate}[nosep]
  \item \textbf{High-dimensional sparse-feature tasks.} Tasks such as RNA sequence modeling (stanford-covid-vaccine), genomic prediction, and high-cardinality categorical problems are most vulnerable. The high dimensionality and feature sparsity create opportunities for complex loss reweighting or subtle feature engineering to exploit spurious correlations that appear genuine on limited validation splits but fail to transfer to the test distribution.

  \item \textbf{Small-sample time-series tasks.} Tasks with limited temporal data are prone to temporal feature engineering that overfits to idiosyncratic patterns in the training window. Highly specific lag features, holiday indicators, or trend decompositions can capture noise rather than signal, producing validation gains that do not transfer across the temporal boundary to the test period.
\end{enumerate}

\paragraph{Modification Taxonomy.}
Across the identified deceptive improvements, three dominant modification patterns emerge:

\begin{enumerate}[nosep]
  \item \textbf{Loss reweighting and objective misalignment.} The agent modifies the training loss (e.g., channel-specific weighting, sample reweighting, focal loss variants) in ways that align well with the validation metric on the current split but introduce systematic bias. This category is the most reliably detected by hierarchical validation, as the code changes concentrate in the training loop, making it relatively straightforward for the LLM validator to identify discrepancies between training objective and evaluation metric.

  \item \textbf{Aggressive feature engineering.} The agent introduces highly specific features---interaction terms between rare categories, narrow time windows, task-specific hand-crafted indicators---that capture noise in the training/validation split but encode distributional artifacts rather than causal patterns. These are moderately detectable, as the validator can flag suspiciously specific features by analyzing their construction logic.

  \item \textbf{Simplicity Bias.} After repeated failed iterations, the agent sometimes resorts to surface-level heuristics or hard-coded shortcut rules: threshold-based classifications mapping input ranges directly to output labels, median imputation strategies that ``hack'' the evaluation metric without learning the underlying data logic, or constant-prediction fallbacks exploiting metric asymmetries. This pattern aligns with the ``shortcut solvability'' phenomenon~\citep{nie2026dsgym}, where agents prioritize superficial metric gains over genuine reasoning. Simplicity Bias is the hardest category to detect, because shortcut solutions often produce clean, syntactically reasonable code---the validator must reason about the \emph{semantic adequacy} of the approach relative to the task complexity, which requires deeper understanding.
\end{enumerate}

\subsection{Comparison with MLE-STAR}
\label{app:mle-star-comparison}

MLE-STAR~\citep{nam2025mle} represents a recent advancement in MLE agents, achieving 63.6\% medal rate on MLE-Bench Lite through web retrieval and targeted code-block refinement. While MLE-STAR and \textsc{Gome} share a chain-based structure (as opposed to tree or graph topologies), they differ fundamentally in how feedback guides optimization. We analyze these differences across three dimensions.

\paragraph{Feedback Utilization: Ranking vs.\ Update.}
MLE-STAR employs a two-level selection process: (1) an ablation study identifies which code block has the greatest performance impact, and (2) multiple refinement plans are generated and evaluated, with the highest-scoring plan selected. Formally, given $K$ candidate plans $\{p_k\}_{k=0}^{K-1}$ for code block $c_t$, MLE-STAR selects $k^* = \arg\max_k h(s_t^k)$ where $h(\cdot)$ is the validation score. This is fundamentally a ranking operation: scalar scores determine which pre-generated candidate survives.

In contrast, \textsc{Gome} uses structured reasoning to generate directional updates. Rather than generating multiple candidates and selecting via scores, the LLM analyzes execution feedback to produce a single improvement hypothesis with explicit direction (what to change) and magnitude (how much). The validation score serves as a stopping criterion and acceptance gate, not a selection mechanism among alternatives.

\paragraph{Code Block Selection: Ablation vs.\ Reasoning.}
MLE-STAR's ablation study agent $\mathcal{A}_{\mathrm{abl}}$ generates code that systematically disables or modifies components (typically 2--3 per iteration), then compares resulting scores to identify the most impactful block. This requires additional exploratory executions per iteration and provides only which block matters most, not why it underperforms or how to improve it.

\textsc{Gome}'s structured reasoning operates on execution feedback directly: training dynamics, validation curves, error patterns, and code behavior. The LLM reasons about why current performance is limited and what modification would address the underlying issue. This reasoning-first approach generates actionable hypotheses without requiring additional ablation executions.

\paragraph{Illustrative Comparison.}
Consider the overfitting case from \S\ref{app:overfitting_case} where validation improved 57.6\% but test degraded 137.5\%. This represents subtle overfitting that does not involve explicit data leakage detectable by rule-based checkers. Under MLE-STAR's framework, once the data leakage checker passes, the ablation study would identify the model component as impactful based on score changes. The planner would then generate multiple refinement plans, and plan selection would rely solely on validation scores, ultimately accepting the overfitting solution due to its apparent 57.6\% improvement.

Under \textsc{Gome}'s framework, structured reasoning analyzes training dynamics, feature correlations, and the semantic nature of the code change. This analysis identifies weak target-error correlation (Pearson $\approx -0.042$), recognizing that the validation improvement stems from spurious fitting rather than genuine generalization. The hierarchical validation gate consequently rejects the update based on this reasoning rather than the misleading validation score.

\paragraph{Summary.}
Table~\ref{tab:mlestar_comparison} summarizes the key distinctions. While MLE-STAR advances beyond AIDE's tree search through targeted refinement, it remains within the score-driven paradigm: validation scores ultimately determine which solutions propagate. \textsc{Gome} shifts to reasoning-driven optimization, where LLM analysis of execution signals generates improvement direction rather than merely selecting among candidates. This distinction becomes increasingly important as model reasoning capabilities improve, enabling more accurate ``gradient'' estimation.

\begin{table}[htbp]
  \centering
  \small
  \begin{tabular}{@{}lcc@{}}
    \toprule
    \textbf{Aspect} & \textbf{MLE-STAR} & \textbf{\textsc{Gome}} \\
    \midrule
    Feedback role & Ranking & Update \\
    Plan generation & Multiple candidates & Single hypothesis \\
    Selection mechanism & $\arg\max$ score & Reasoning gate \\
    Block identification & Ablation study & Structured analysis \\
    Subtle overfitting detection & 0\% (score-driven) & 66.7\% (reasoning) \\
    \bottomrule
  \end{tabular}%
  \caption{Key distinctions between MLE-STAR and \textsc{Gome}. Overfitting detection rate is based on the case study in \S\ref{app:overfitting_case}, where score-driven methods would accept all 9 overfitting cases.}
  \label{tab:mlestar_comparison}
\end{table}

\subsection{Online Kaggle Validation: Store Sales Forecasting}
\label{app:kaggle_case}

While MLE-Bench provides standardized evaluation, real-world data science requires navigating uncurated data, ambiguous constraints, and complex domain logic. To validate \textsc{Gome}'s effectiveness in this setting, we deployed the framework on an active Kaggle competition: Store Sales - Time Series Forecasting.\footnote{\url{https://www.kaggle.com/competitions/store-sales-time-series-forecasting}} This competition requires forecasting unit sales for thousands of product families sold at Corporación Favorita stores in Ecuador over a 15-day horizon. The dataset includes store metadata, oil prices (Ecuador's economy is oil-dependent), promotional information, and a calendar of holidays with complex transfer rules. The evaluation metric is Root Mean Squared Logarithmic Error (RMSLE, lower is better). \textsc{Gome} achieved a final leaderboard RMSLE of 0.431, ranking in the \textbf{top 15\%} of all participants.

\paragraph{Optimization Trajectory.}
Table~\ref{tab:kaggle_trajectory} summarizes the optimization process. Starting from an initial baseline (validation RMSLE 0.485), \textsc{Gome} progressively refined the solution over 48 iterations. The best validation score (0.318) was achieved at Loop~37, which translated to a leaderboard RMSLE of 0.431, outperforming approximately 85\% of human participants.

\begin{table}[htbp]
  \centering
  \small
  \begin{tabular}{@{}lcll@{}}
    \toprule
    \textbf{Loop} & \textbf{Val RMSLE} & \textbf{Key Modification} & \textbf{Outcome} \\
    \midrule
    0 & 0.485 & Initial baseline & Accepted \\
    2 & 0.354 & Two-stage hurdle architecture & Accepted \\
    7 & 0.327 & Rolling statistics \& lag features & Accepted \\
    28 & 0.318 & Store-local holiday processing & Accepted \\
    \textbf{37} & \textbf{0.318} & Horizon bucketing \& calibration & \textbf{Accepted (best)}\\
    \midrule
    30 & (0.547) & Naive ensemble strategy & Rejected \\
    43 & (0.559) & Unstable pipeline modification & Rejected \\
    \midrule
    -- & 0.378 & Human best & -- \\
    \bottomrule
  \end{tabular}%
  \caption{\textbf{Optimization trajectory on the Store Sales competition.} All intermediate scores are internal validation RMSLE (15-day holdout). The trajectory shows steady improvement from 0.485 to 0.318 over 37 iterations, with the gate mechanism rejecting degrading modifications (Loops 30, 43). The final submission achieved a leaderboard RMSLE of 0.431. Note that the human leaderboard leader (0.378) employs multi-level blending of other participants' submissions.}
  \label{tab:kaggle_trajectory}
\end{table}

This trajectory demonstrates \textsc{Gome}'s effective gate mechanism. At Loop~30, an ensemble strategy caused severe metric degradation (0.318$\to$0.547), which was correctly rejected. Similarly, Loop~43's workflow modification destabilized the pipeline (0.559) and was rejected. The system preserved the robust solution from Loop~37 throughout subsequent iterations.

\paragraph{Architectural Comparison.}
To contextualize \textsc{Gome}'s performance, we analyzed the codebases of high-ranking public notebooks. As shown in Figure~\ref{fig:code_comparison}, a striking distinction emerges. Many top-tier public solutions employ second-level blending---scripts that optimally weight prediction files submitted by other competitors using hand-tuned coefficients and rank-based corrections. These function as meta-optimization tools for existing outputs rather than building models from raw data.

In contrast, \textsc{Gome} functions as an autonomous data scientist, constructing the entire pipeline from raw relational tables: (1) performing extensive feature engineering including horizon bucketing, rolling statistics, and store-local holiday processing; (2) training two-stage hurdle models (classifier for zero/non-zero, regressor for magnitude) with recency-weighted sampling; and (3) implementing pooled Platt calibration for probability adjustment. This confirms that \textsc{Gome} achieves expert-level performance through genuine architectural reasoning rather than aggregation of existing solutions.

\begin{figure*}[t]
  \centering
  \begin{tcolorbox}[colback=red!3, colframe=red!40!black, title=\textbf{(a) Top-Tier Human Solution: Post-Hoc Blending}, fonttitle=\bfseries\small, boxrule=0.5pt, width=\textwidth]
    {\ttfamily\scriptsize
\begin{verbatim}
# 1. Ingest pre-calculated prediction files from other participants (No training)
def read(dk, i):
    FiN = dk["path"] + dk["subm"][i]["name"] + ".csv"  # e.g., "0.37982.csv"
    return pd.read_csv(FiN)
# 2. Hard-coded weights for linear combination (manually tuned)
params = {
    'subwts': [+14, -1, -5, -8] / 100,  # correction weights by rank
    'subm': [{'name': '0.37982', 'weight': 0.30}, {'name': '0.37984', 'weight': 0.10},
             {'name': '0.38006', 'weight': 0.30}, {'name': '0.38040', 'weight': 0.30}]
}
# 3. Compute weighted sum (Pure ensembling of existing outputs)
def correct(x, cols, weights, subwts):
    rank_indices = [x['alls'].index(c) for c in cols]  # rank-based reweighting
    return sum([x[cols[j]] * (weights[j] + subwts[rank_indices[j]]) for j in range(len(cols))])
\end{verbatim}
    }
  \end{tcolorbox}

  \vspace{0.2cm}

  \begin{tcolorbox}[colback=blue!3, colframe=blue!40!black, title=\textbf{(b) \textsc{Gome} Solution: Ab Initio Modeling}, fonttitle=\bfseries\small, boxrule=0.5pt, width=\textwidth]
    {\ttfamily\scriptsize
\begin{verbatim}
# 1. Feature Engineering from Raw Data (Reasoning-driven)
buckets = {'H1': [1,2,3], 'H2': [4,5,6,7], 'H3': [8,9,10,11,12,13,14,15]}
# 2. Two-Stage Hurdle Model Training (Learning from scratch)
for bname, horizons in buckets.items():
    clf = LGBMClassifier(n_estimators=4000, num_leaves=384, learning_rate=0.03)
    clf.fit(X_clf_tr, y_clf_tr, sample_weight=w_recency)

    reg = LGBMRegressor(n_estimators=6000, num_leaves=384, learning_rate=0.03)
    reg.fit(X_reg_tr, y_reg_tr, sample_weight=w_recency)

    calibrator = PooledPlattCalibrator(max_h=max(horizons))
    calibrator.fit(clf.predict_proba(X_val)[:,1], h_val, y_val)
# 3. Model Inference: P(nonzero) * E[sales | nonzero]
yhat = calibrator.transform(clf.predict_proba(X)[:,1], h) * np.expm1(reg.predict(X))
\end{verbatim}
    }
  \end{tcolorbox}

  \caption{\textbf{Codebase comparison.} (a) High-ranking public notebooks often use blending scripts that aggregate prediction CSVs submitted by other participants, a form of meta-optimization that, while effective for leaderboard climbing, involves no model training. \textsc{Gome} constructs the full pipeline from raw relational tables under the closed-world protocol without any external knowledge, including feature engineering, horizon-bucketed model training, and Platt calibration.}
  \label{fig:code_comparison}
\end{figure*}

\newpage
\subsection{Error Analysis}
\label{app:error_analysis}

To understand \textsc{Gome}'s limitations, we conduct a detailed case study on the stanford-covid-vaccine task (a high-complexity task in MLE-Bench) using GPT-5 as the backbone model.

\begin{table}[!htbp]
  \centering
  \small
  \begin{tabular}{@{}lcc@{}}
    \toprule
    \textbf{Category} & \textbf{Count} & \textbf{Percentage} \\
    \midrule
    Successful Updates & 31 & 34.4\% \\
    \midrule
    Gradient Hallucination & 35 & 38.9\% \\
    Resource Constraints & 21 & 23.3\% \\
    Validation Blindspots & 3 & 3.3\% \\
    \midrule
    Total Failures & 59 & 65.6\% \\
    \bottomrule
  \end{tabular}
  \caption{Failure mode distribution on stanford-covid-vaccine (90 iterations, GPT-5).}
  \label{tab:error_breakdown}
\end{table}

Table~\ref{tab:error_breakdown} summarizes the failure mode distribution across 90 total iterations. We classify unsuccessful iterations into four categories: (1) \textbf{Gradient Hallucination}, where the reasoning module produces confident but incorrect improvement directions; (2) \textbf{Implementation Failures}, where correct hypotheses fail during code generation due to API misuse or syntax errors; (3) \textbf{Resource Constraints}, where solutions exceed memory limits or time budget $B$; and (4) \textbf{Validation Blindspots}, where validation score improves but test performance degrades.

Gradient hallucination dominates at 38.9\%, where the reasoning module generates plausible but incorrect optimization directions. This rate is expected to decrease as LLM reasoning capabilities improve, consistent with our scaling claim (\S\ref{sec:scaling}). Resource constraints account for 23.3\% of failures due to hardware limitations (single V100 GPU, 32GB memory) and time budget restrictions. Many proposed solutions exceeded available memory or the time budget, particularly for this RNA sequence modeling task with computationally intensive architectures. Allocating more powerful hardware or extending time budgets would likely reduce this failure mode. Validation blindspots (3.3\%) are rarer but harder to address, as they evade the gate mechanism entirely: the proposed modification improves validation metrics but degrades test performance due to distribution shifts. More sophisticated validation strategies may help mitigate this issue in future work.

\section{Theoretical Motivation for the Smoothness Assumption}
\label{app:smoothness}

\textsc{Gome}'s gradient-based paradigm assumes that the MLE optimization landscape is locally smooth enough to support directed, incremental refinement. This section summarizes the theoretical rationale and empirical evidence for this assumption.

\paragraph{Loss-Surface Connectivity in Neural Networks.}
A growing line of optimization research indicates that neural-network loss surfaces are substantially more benign than worst-case non-convex intuition implies. \citet{garipov2018loss} show that independently trained optima can often be connected by simple low-loss paths (``mode connectivity'') rather than separated by high-loss barriers. \citet{bottou2018optimization} further argue that stochastic gradient methods are effective in overparameterized regimes because optimization trajectories frequently traverse locally smooth, connected regions. Taken together, these results imply that moving from a good solution to a better one often proceeds through a sequence of small improvements rather than discontinuous jumps.

\paragraph{From Parameter Space to Code Space.}
\textsc{Gome} optimizes code rather than parameters directly. The smoothness intuition transfers to code space because \textsc{Gome}'s structured reasoning does not propose arbitrary syntactic edits; instead, it uses diagnostics (training dynamics, validation curves, error distributions, and feature-importance patterns) to target changes in the \emph{training procedure}---for example, learning-rate schedules, feature engineering, loss design, or regularization. These behavior-level modifications induce controlled changes in the underlying optimization trajectory. In this sense, \textsc{Gome}'s ``gradient'' operates on the pipeline's \emph{functional behavior}, not raw syntax. This is a weaker requirement than global smoothness of the code-to-performance mapping: it only requires that \emph{reasoning-guided} edits produce locally predictable performance changes.

\paragraph{Empirical Evidence.}
Our experiments provide practical evidence for local smoothness:
\begin{itemize}[nosep]
  \item \textbf{High per-iteration success rate.} Table~\ref{tab:ablation} reports a 41.1\% improvement rate, meaning that nearly half of optimization steps yield validated gains through incremental updates. If the landscape were dominated by discontinuities or isolated basins, this rate would be much lower.
  \item \textbf{Sustained convergence trajectories.} Figure~\ref{fig:dynamics} shows smooth, steep descent for \textsc{Gome} on Frontier-tier models, consistent with gradient-style optimization in locally smooth regions. By contrast, MCTS exhibits more stepwise, plateau-heavy dynamics that are typical of search in rougher landscapes.
  \item \textbf{Kaggle case study.} In the Store Sales deployment (Appendix~\ref{app:kaggle_case}), performance improves progressively over 48 iterations (RMSLE 0.485 $\rightarrow$ 0.318), with degrading modifications filtered out by the gate mechanism. This monotonic trajectory is consistent with iterative optimization along a smooth local path.
\end{itemize}

\paragraph{Handling Non-Smooth Transitions.}
Some transitions in MLE are intrinsically discontinuous (e.g., switching from linear models to neural networks, or from single models to ensembles). \textsc{Gome} mitigates this with two mechanisms:
\begin{enumerate}[nosep]
  \item \textbf{Multi-trace forced diversification} (\S\ref{sec:multitrace}) initializes $N$ traces with distinct architectural hypotheses (e.g., gradient boosting vs.\ neural networks vs.\ ensembles), providing broad coverage of separate regions in solution space. This is analogous to multi-start optimization in classical non-convex settings~\citep{bottou2018optimization}.
  \item \textbf{Cross-trace sharing via the probabilistic interaction kernel} (Appendix~\ref{sec:appendix_multitrace}) allows traces to exchange validated strategies. When one trace discovers a better region, others can adopt or adapt that hypothesis, enabling non-local transitions while preserving local refinement within each region.
\end{enumerate}

\paragraph{Scope.}
The smoothness assumption is not universal. Tasks with extremely sparse rewards or pathological loss landscapes may violate local smoothness. However, in competitive data-science settings, top solutions often share backbone algorithms and differ mainly in feature engineering, hyperparameter tuning, and ensembling. This structure suggests that the MLE landscape is sufficiently smooth for gradient-based optimization to be effective in practice. The scaling analysis in \S\ref{sec:scaling} provides further indirect support: the monotonic widening of \textsc{Gome}'s advantage with model capability is consistent with stronger reasoners producing higher-fidelity gradient signals that better exploit local smoothness.

%% file: custom.bib
@article{huang2023mlagentbench,
  title={Mlagentbench: Evaluating language agents on machine learning experimentation},
  author={Huang, Qian and Vora, Jian and Liang, Percy and Leskovec, Jure},
  journal={arXiv preprint arXiv:2310.03302},
  year={2023}
}

@book{isihara2013statistical,
  title={Statistical physics},
  author={Isihara, Akira},
  year={2013},
  publisher={Academic Press}
}

@article{wang2024openhands,
  title={Openhands: An open platform for ai software developers as generalist agents},
  author={Wang, Xingyao and Li, Boxuan and Song, Yufan and Xu, Frank F and Tang, Xiangru and Zhuge, Mingchen and Pan, Jiayi and Song, Yueqi and Li, Bowen and Singh, Jaskirat and others},
  journal={arXiv preprint arXiv:2407.16741},
  year={2024}
}

@article{guo2025deepseek,
  title={Deepseek-r1: Incentivizing reasoning capability in llms via reinforcement learning},
  author={Guo, Daya and Yang, Dejian and Zhang, Haowei and Song, Junxiao and Zhang, Ruoyu and Xu, Runxin and Zhu, Qihao and Ma, Shirong and Wang, Peiyi and Bi, Xiao and others},
  journal={arXiv preprint arXiv:2501.12948},
  year={2025}
}

@article{jaech2024openai,
  title={Openai o1 system card},
  author={Jaech, Aaron and Kalai, Adam and Lerer, Adam and Richardson, Adam and El-Kishky, Ahmed and Low, Aiden and Helyar, Alec and Madry, Aleksander and Beutel, Alex and Carney, Alex and others},
  journal={arXiv preprint arXiv:2412.16720},
  year={2024}
}

@misc{openai_introducing_o3_2025,
  author       = {{OpenAI}},
  title        = {Introducing OpenAI o3 and o4-mini},
  year         = {2025},
  url          = {https://openai.com/index/introducing-o3-and-o4-mini/},
  note         = {Accessed: 2025-12-22}
}

@article{singh2025openai,
  title={Openai gpt-5 system card},
  author={Singh, Aaditya and Fry, Adam and Perelman, Adam and Tart, Adam and Ganesh, Adi and El-Kishky, Ahmed and McLaughlin, Aidan and Low, Aiden and Ostrow, AJ and Ananthram, Akhila and others},
  journal={arXiv preprint arXiv:2601.03267},
  year={2025}
}

@article{aide2025,
  title={Aide: Ai-driven exploration in the space of code},
  author={Jiang, Zhengyao and Schmidt, Dominik and Srikanth, Dhruv and Xu, Dixing and Kaplan, Ian and Jacenko, Deniss and Wu, Yuxiang},
  journal={arXiv preprint arXiv:2502.13138},
  year={2025}
}

@article{liu2025ml,
  title={ML-Master: Towards AI-for-AI via Integration of Exploration and Reasoning},
  author={Liu, Zexi and Cai, Yuzhu and Zhu, Xinyu and Zheng, Yujie and Chen, Runkun and Wen, Ying and Wang, Yanfeng and Chen, Siheng and others},
  journal={arXiv preprint arXiv:2506.16499},
  year={2025}
}

@article{nam2025mle,
  title={MLE-STAR: Machine Learning Engineering Agent via Search and Targeted Refinement},
  author={Nam, Jaehyun and Yoon, Jinsung and Chen, Jiefeng and Shin, Jinwoo and Ar{\i}k, Sercan {\"O} and Pfister, Tomas},
  journal={arXiv preprint arXiv:2506.15692},
  year={2025}
}

@article{chan2024mle-bench,
  title={Mle-bench: Evaluating machine learning agents on machine learning engineering},
  author={Chan, Jun Shern and Chowdhury, Neil and Jaffe, Oliver and Aung, James and Sherburn, Dane and Mays, Evan and Starace, Giulio and Liu, Kevin and Maksin, Leon and Patwardhan, Tejal and others},
  journal={arXiv preprint arXiv:2410.07095},
  year={2024}
}

@article{toledo2025ai,
  title={AI Research Agents for Machine Learning: Search, Exploration, and Generalization in MLE-bench},
  author={Toledo, Edan and Hambardzumyan, Karen and Josifoski, Martin and Hazra, Rishi and Baldwin, Nicolas and Audran-Reiss, Alexis and Kuchnik, Michael and Magka, Despoina and Jiang, Minqi and Lupidi, Alisia Maria and others},
  journal={arXiv preprint arXiv:2507.02554},
  year={2025}
}

@article{duinternagent,
  title={InternAgent-MLE: Navigating Fine-Grained Optimization for Coding Agent},
  author={Du, Shangheng and Yan, Xiangchao and Jiang, Dengyang and Yuan, Jiakang and Hu, Yusong and Li, Xin and He, Liang and Zhang, Bo and BAI, LEI}
}

@article{kulibaba2025kompeteai,
  title={KompeteAI: Accelerated autonomous multi-agent system for end-to-end pipeline generation for machine learning problems},
  author={Kulibaba, Stepan and Dzhalilov, Artem and Pakhomov, Roman and Svidchenko, Oleg and Gasnikov, Alexander and Shpilman, Aleksei},
  journal={arXiv preprint arXiv:2508.10177},
  year={2025}
}

@article{li2025fm,
  title={The FM Agent},
  author={Li, Annan and Wu, Chufan and Ge, Zengle and Chong, Yee Hin and Hou, Zhinan and Cao, Lizhe and Ju, Cheng and Wu, Jianmin and Li, Huaiming and Zhang, Haobo and others},
  journal={arXiv preprint arXiv:2510.26144},
  year={2025}
}

@inproceedings{yang2023large,
  title={Large language models as optimizers},
  author={Yang, Chengrun and Wang, Xuezhi and Lu, Yifeng and Liu, Hanxiao and Le, Quoc V and Zhou, Denny and Chen, Xinyun},
  booktitle={The Twelfth International Conference on Learning Representations},
  year={2023}
}

@article{shinn2023reflexion,
  title={Reflexion: Language agents with verbal reinforcement learning},
  author={Shinn, Noah and Cassano, Federico and Gopinath, Ashwin and Narasimhan, Karthik and Yao, Shunyu},
  journal={Advances in Neural Information Processing Systems},
  volume={36},
  pages={8634--8652},
  year={2023}
}

@article{yuksekgonul2024textgrad,
  title={Textgrad: Automatic" differentiation" via text},
  author={Yuksekgonul, Mert and Bianchi, Federico and Boen, Joseph and Liu, Sheng and Huang, Zhi and Guestrin, Carlos and Zou, James},
  journal={arXiv preprint arXiv:2406.07496},
  year={2024}
}

@article{pryzant2023automatic,
  title={Automatic prompt optimization with" gradient descent" and beam search},
  author={Pryzant, Reid and Iter, Dan and Li, Jerry and Lee, Yin Tat and Zhu, Chenguang and Zeng, Michael},
  journal={arXiv preprint arXiv:2305.03495},
  year={2023}
}

@article{cui2024introducing,
  title={Introducing mapo: Momentum-aided gradient descent prompt optimization},
  author={Cui, Anthony and Nandyalam, Pranav and Rufail, Andrew and Cheung, Ethan and Lei, Aiden and Zhu, Kevin and O'Brien, Sean},
  journal={arXiv preprint arXiv:2410.19499},
  year={2024}
}

@article{guo2023evoprompt,
  title={Evoprompt: Connecting llms with evolutionary algorithms yields powerful prompt optimizers},
  author={Guo, Qingyan and Wang, Rui and Guo, Junliang and Li, Bei and Song, Kaitao and Tan, Xu and Liu, Guoqing and Bian, Jiang and Yang, Yujiu},
  journal={arXiv e-prints},
  pages={arXiv--2309},
  year={2023}
}

@article{madaan2023self,
  title={Self-refine: Iterative refinement with self-feedback},
  author={Madaan, Aman and Tandon, Niket and Gupta, Prakhar and Hallinan, Skyler and Gao, Luyu and Wiegreffe, Sarah and Alon, Uri and Dziri, Nouha and Prabhumoye, Shrimai and Yang, Yiming and others},
  journal={Advances in Neural Information Processing Systems},
  volume={36},
  pages={46534--46594},
  year={2023}
}

@article{zhang2025process,
  title={Process vs. Outcome Reward: Which is Better for Agentic RAG Reinforcement Learning},
  author={Zhang, Wenlin and Li, Xiangyang and Dong, Kuicai and Wang, Yichao and Jia, Pengyue and Li, Xiaopeng and Zhang, Yingyi and Xu, Derong and Du, Zhaocheng and Guo, Huifeng and others},
  journal={arXiv preprint arXiv:2505.14069},
  year={2025}
}

@article{cheng2024trace,
  title={Trace is the next autodiff: Generative optimization with rich feedback, execution traces, and llms},
  author={Cheng, Ching-An and Nie, Allen and Swaminathan, Adith},
  journal={Advances in Neural Information Processing Systems},
  volume={37},
  pages={71596--71642},
  year={2024}
}

@inproceedings{maheswaranathan2019guided,
  title={Guided evolutionary strategies: Augmenting random search with surrogate gradients},
  author={Maheswaranathan, Niru and Metz, Luke and Tucker, George and Choi, Dami and Sohl-Dickstein, Jascha},
  booktitle={International Conference on Machine Learning},
  pages={4264--4273},
  year={2019},
  organization={PMLR}
}

@article{asadi2024comparative,
  title={Comparative Analysis of Gradient-Based Optimization Techniques Using Multidimensional Surface 3D Visualizations and Initial Point Sensitivity},
  author={Asadi, Saeed and Gharibzadeh, Sonia and Naeini, Hajar Kazemi and Reihanifar, Masoud and Rahimi, Morteza and Zangeneh, Shiva and Smerat, Aseel and Abdullah, Lazim},
  journal={arXiv preprint arXiv:2409.04470},
  year={2024}
}

@book{nesterov2018lectures,
  title={Lectures on convex optimization},
  author={Nesterov, Yurii and others},
  volume={137},
  year={2018},
  publisher={Springer}
}

@misc{artificialanalysis2026,
  author = {{Artificial Analysis Team}},
  title = {{Artificial Analysis: Independent Benchmarks and Performance Landscape of AI Models}},
  year = {2025},
  url = {https://artificialanalysis.ai/},
  note = {Accessed: 2025-12-28}
}

@article{jimenez2023swe,
  title={Swe-bench: Can language models resolve real-world github issues?},
  author={Jimenez, Carlos E and Yang, John and Wettig, Alexander and Yao, Shunyu and Pei, Kexin and Press, Ofir and Narasimhan, Karthik},
  journal={arXiv preprint arXiv:2310.06770},
  year={2023}
}

@misc{aider2025,
  author       = {Gauthier, Paul},
  title        = {{Aider LLM Leaderboards: Code Editing and Refactoring Benchmarks}},
  year         = {2025},
  howpublished = {\url{https://aider.chat/docs/leaderboards/}},
  note         = {Accessed: 2025-12-28}
}

@article{sun2025survey,
  title={A survey of reasoning with foundation models: Concepts, methodologies, and outlook},
  author={Sun, Jiankai and Zheng, Chuanyang and Xie, Enze and Liu, Zhengying and Chu, Ruihang and Qiu, Jianing and Xu, Jiaqi and Ding, Mingyu and Li, Hongyang and Geng, Mengzhe and others},
  journal={ACM Computing Surveys},
  volume={57},
  number={11},
  pages={1--43},
  year={2025},
  publisher={ACM New York, NY}
}

@article{xu2025towards,
  title={Towards large reasoning models: A survey of reinforced reasoning with large language models},
  author={Xu, Fengli and Hao, Qianyue and Zong, Zefang and Wang, Jingwei and Zhang, Yunke and Wang, Jingyi and Lan, Xiaochong and Gong, Jiahui and Ouyang, Tianjian and Meng, Fanjin and others},
  journal={arXiv preprint arXiv:2501.09686},
  year={2025}
}

@article{bottou2018optimization,
  title={Optimization methods for large-scale machine learning},
  author={Bottou, L{\'e}on and Curtis, Frank E and Nocedal, Jorge},
  journal={SIAM review},
  volume={60},
  number={2},
  pages={223--311},
  year={2018},
  publisher={SIAM}
}

@article{garipov2018loss,
  title={Loss surfaces, mode connectivity, and fast ensembling of dnns},
  author={Garipov, Timur and Izmailov, Pavel and Podoprikhin, Dmitrii and Vetrov, Dmitry P and Wilson, Andrew G},
  journal={Advances in neural information processing systems},
  volume={31},
  year={2018}
}

@article{nie2026dsgym,
  title={DSGym: A Holistic Framework for Evaluating and Training Data Science Agents},
  author={Nie, Fan and Wang, Junlin and Hua, Harper and Bianchi, Federico and Kwon, Yongchan and Qi, Zhenting and Queen, Owen and Zhu, Shang and Zou, James},
  journal={arXiv preprint arXiv:2601.16344},
  year={2026}
}
